\documentclass[10pt,twocolumn,letterpaper]{article}

\usepackage{cvpr}              
\usepackage{amsmath}
\usepackage{multirow}
\usepackage{algorithm}
\usepackage{algorithmic}
\usepackage[table]{xcolor}
\usepackage[accsupp]{axessibility}
\definecolor{gray}{rgb}{0.929, 0.929, 0.929}
\definecolor{blue2}{rgb}{0.9098, 0.9608, 0.9882}
\definecolor{blue3}{rgb}{0.7598, 0.8108, 0.8382}
\definecolor{cvprblue}{rgb}{0.21,0.49,0.74}
\usepackage[pagebackref,breaklinks,colorlinks,allcolors=cvprblue]{hyperref}

\def\paperID{***} 
\def\confName{CVPR}
\def\confYear{2026}

\title{Occlusion-Aware SORT: Observing Occlusion for Robust Multi-Object Tracking}

\author{
Chunjiang Li\textsuperscript{1} \quad
Jianbo Ma\textsuperscript{2} \quad
Li Shen\textsuperscript{1} \quad
Yanru Chen\textsuperscript{1}\thanks{Corresponding author} \quad
Liangyin Chen\textsuperscript{1} \\
\textsuperscript{1}College of Computer Science, Sichuan University, Chengdu, China \\
\textsuperscript{2}Institute of Optics and Electronics, CAS, Chengdu, China \\
{\tt\small lichunjiang@stu.scu.edu.cn, \{chenyanru, chenliangyin\}@scu.edu.cn}
}

\begin{document}
\maketitle
\begin{abstract}
Multi-object tracking (MOT) involves analyzing object trajectories and counting the number of objects in video sequences. However, 2D MOT faces challenges due to positional cost confusion arising from partial occlusion. To address this issue, we present the novel Occlusion-Aware SORT (OA-SORT) framework, a plug-and-play and training-free framework that includes the Occlusion-Aware Module (OAM), the Occlusion-Aware Offset (OAO), and the Bias-Aware Momentum (BAM). Specifically, OAM analyzes the occlusion status of objects, where a Gaussian Map (GM) is introduced to reduce background influence. In contrast, OAO and BAM leverage the OAM-described occlusion status to mitigate cost confusion and suppress estimation instability. Comprehensive evaluations on the DanceTrack, SportsMOT, and MOT17 datasets demonstrate the importance of occlusion handling in MOT. On the DanceTrack test set, OA-SORT achieves 63.1\% and 64.2\% in HOTA and IDF1, respectively. Furthermore, integrating the Occlusion-Aware framework into the four additional trackers improves HOTA and IDF1 by an average of 2.08\% and 3.05\%, demonstrating the reusability of the occlusion awareness.
\end{abstract}

\section{Introduction}
\label{sec: intro}

Multi-Object Tracking (MOT) is used to capture the trajectories of objects in video footage or webcam streams. The fundamental objective is to assign a unique identifier to each detected object \cite{LUO2021103448}. The field has a variety of applications, including the analysis of object movements \cite{9760217}, the tracking of changes in human postures \cite{9954214}, and the accurate counting of objects \cite{TrafficFlow}. Despite significant progress in this area, MOT still faces challenges brought by occlusion.


\begin{figure}[t!]
\centering
\includegraphics[width=\linewidth]{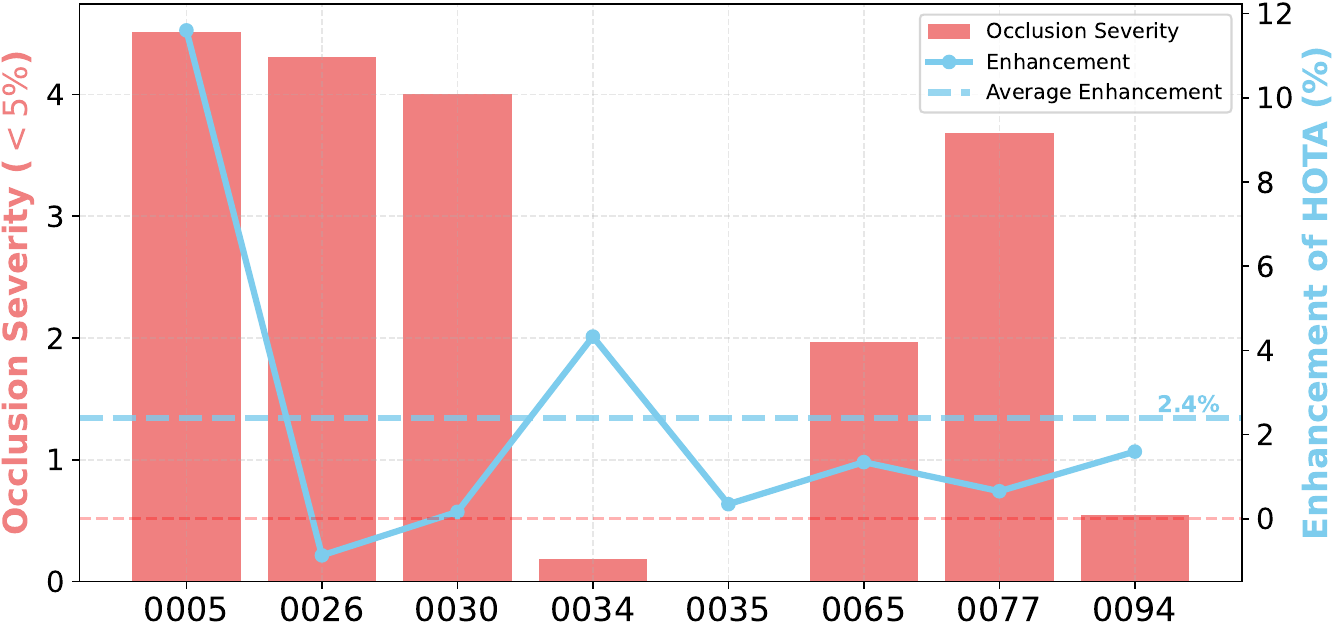}  \\
\includegraphics[width=\linewidth]{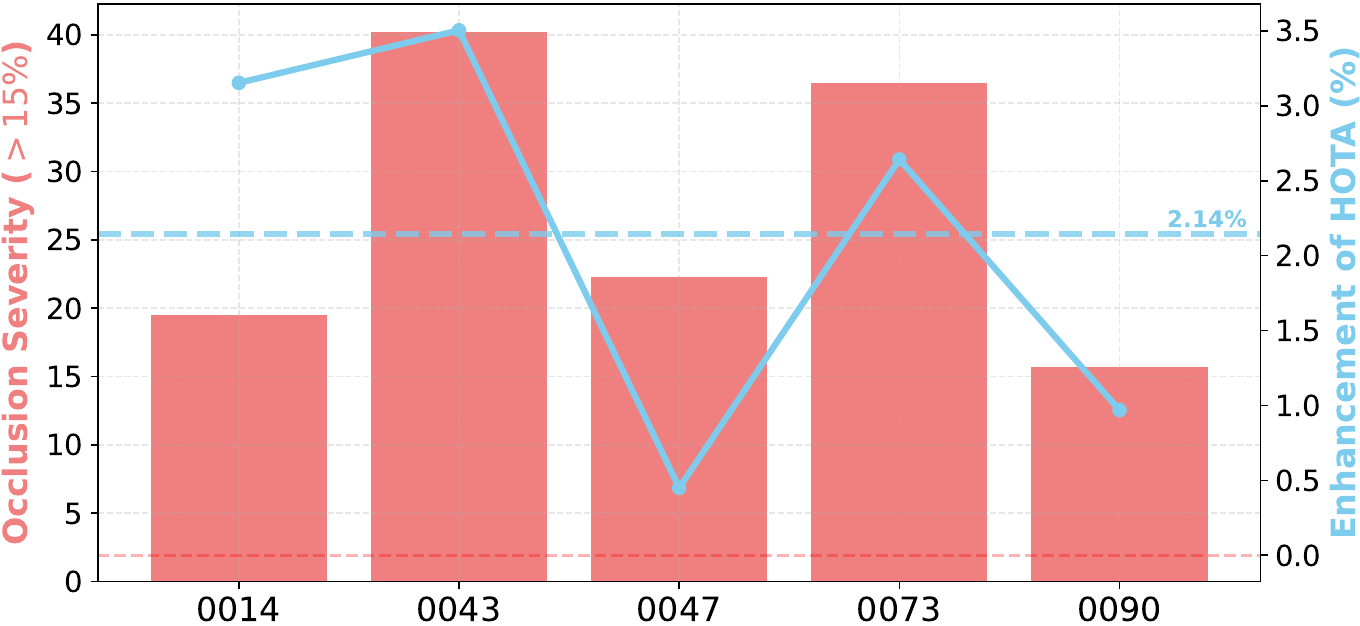}
\caption{Performance improvement of OA-SORT over baseline (Hybrid-SORT) on the DanceTrack validation set under different occlusion severity, where the horizontal axis represents the video sequence number. The severity is defined as the proportion of object instances whose average occlusion ratio (Eq.~\ref{eq: Oc_2}) exceeds 0.75.}
\label{fig:occlusion}
\end{figure}

Currently, position association (widely used in MOT) remains a pivotal element for tracking objects across successive frames. The majority of trackers~\cite{sort, breitenstein2009robust, wang2025pd, zhang2022bytetrack} rely on position association, which utilizes motion cues such as velocity and acceleration to predict the objects' positions in a video sequence. The position association consists of three main components: position predictor~\cite{kalman, GIAOTracker}, spatial consistency metrics \cite{yu2016unitbox, yang2024hybrid}, and the Hungarian algorithm~\cite{kuhn1955hungarian}. However, the tracking process often presents challenges due to partial occlusion. In terms of detection, it becomes difficult to differentiate between the foreground (\ie, the object itself) and the background when objects of the same category exhibit occlusion behavior. This ultimately leads to inaccurate detection. Moreover, discrete and linear position predictors (such as Kalman Filter~\cite{kalman}, KF) are severely affected by frequent inaccurate detections. The impact is more pronounced when the motion of an object is irregular and variable, such as during non-linear posture changes. In the subsequent process of data association between detection and trajectory estimation, unstable predictions and inaccurate detections can easily lead to serious cost confusion. Cost confusion refers to the ambiguity in the generated position cost matrix, which cannot accurately and stably reflect the affinity between detection and estimation. This often leads to frequent or even permanent identity exchanges.

To alleviate this impact, many methods have introduced additional cues, such as appearance features~\cite{deepsort, jde, fairmot, aharon2022bot}, motion direction~\cite{cao2023observation, maggiolino2023deep}, and detection confidence~\cite{zhang2022bytetrack, kang2022rt}. These additional cues improve association performance to some extent. However, inaccurate detections caused by occlusion often compromise the reliability of additional cues. In terms of appearance features, the features' reliability decreases due to inaccurate detections, leading to feature cost confusion. The feature of an occluded object is easily influenced or even replaced by the objects in front of it. Secondly, while motion direction can effectively reduce failed matches during occlusion, cost confusion persists. Similarly, confidence is sensitive to occlusion, and when an object occludes another, the score of the occluded object is affected. In conclusion, while these additional cues enhance association accuracy to a certain extent, cost confusion due to occlusion remains a prevalent issue.

\begin{figure*}[t!]
\centering
\includegraphics[width=\linewidth]{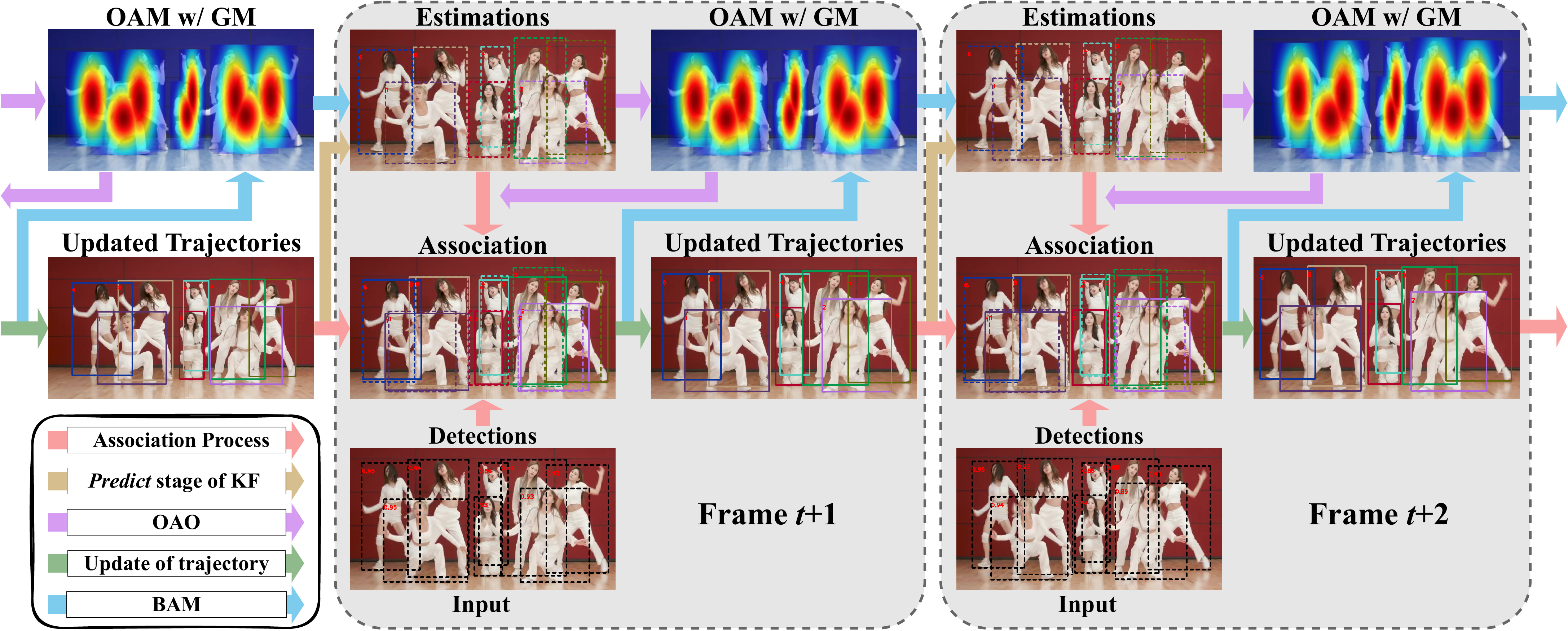}
\caption{Pipeline of OA-SORT (\cref{sec: app}). Before association, OAO (\cref{sec: OAO}) leverages the occlusion coefficient, which is derived from KF's estimation by OAM (\cref{sec: OAM}) with GM (\cref{sec: GM}), to refine the positional cost between trajectories and high-confidence detections. During the update stage, BAM (\cref{sec: BAM}) integrates the occlusion coefficient from the latest observation with IoU to refine the KF update.}
\label{fig: pipeline}
\end{figure*}

To this end, the paper focuses on mitigating positional cost confusion arising from occlusion and introduce a novel approach to observing the occlusion status of objects and employing this to form an occlusion-aware tracking framework. Firstly, the \textit{Occlusion-Aware Module (OAM)} is designed to observe and estimate the occlusion status of objects, \ie, obtaining the occlusion coefficient that reflects occlusion severity. During the calculation process, the \textit{Gauss map (GM)} is introduced to refine the coefficient. On this basis, two new components are proposed, \ie, \textit{Occlusion-Aware Offset (OAO)} and \textit{Bias-Aware Momentum (BAM)}. Finally, a novel occlusion-aware tracking framework, \textit{Occlusion-Aware SORT (OA-SORT)}, is formed by combining these components as shown in Fig.~\ref{fig: pipeline}. Specifically, OAO integrates the occlusion coefficient into the spatial consistency metric to mitigate cost confusion, where the coefficient is used to describe the estimations from the position predictor. BAM aims to reduce the impact caused by inaccurate detections by combining the occlusion coefficient with the spatial consistency metric, thereby enhancing robustness of the position predictor. The occlusion coefficient is used to describe the latest observation of trajectories. Finally, on the DanceTrack, OA-SORT achieves 63.1\%, 48.5\%, and 64.2\% in HOTA, AssA, and IDF1. Our contributions are summarized as follows:
\begin{itemize}
\item The presented OAM observes and evaluates the occlusion status, and GM is introduced to reduce background interference.
\item Based on the OAM, the designed OAO and BAM mitigate cost confusion and optimize the predictor's estimations, demonstrating that integrating occlusion status effectively reduces occlusion-induced issues. Building on those, we design the occlusion-aware framework, OA-SORT, achieving outstanding performance across various occlusion scenarios, as shown in \cref{fig:occlusion}.
\item OAM, OAO, BAM, and occlusion-aware framework can be easily integrated into diverse association strategies and architectures. Experiments on DanceTrack show that occlusion-aware framework brings benefits for the four association methods \cite{zhang2022bytetrack, cao2023observation, liu2025sparsetrack, wang2025pd} with improvements of average +2.08\% HOTA and +3.05\% IDF1. Further ablation studies, including other trackers~\cite{sort, zhang2022bytetrack, cao2023observation, shim2025focusing}, on the DanceTrack validation set verify the generalization and effectiveness of OAO and BAM. The rationality of the occlusion-aware framework is fully explained.
\end{itemize}
\section{Related work}\label{sec: 2}
Association \cite{Jiaqi, sort, zhang2022bytetrack, bochinski2017high, qin2023motiontrack} aims to match identical objects across frames. For association based on position information, two primary paradigms are commonly employed.

\noindent \textbf{Position-Association} relies exclusively on positional information to associate trajectories with detected objects. To address the non-linearity limitations of KF, some works propose improved prediction methods, such as the Unscented KF \cite{julier1997new}, Extended KF \cite{kim2020extended}, and NSA KF \cite{GIAOTracker}. Although these extensions improve motion modeling accuracy, they still rely on KF's motion estimates. Alternatively, some methods \cite{Centertrack, qin2023motiontrack} integrate learnable models to capture non-linear motion factors. However, occlusion remains an unresolved problem in the field. To address this challenge, some methods \cite{Stadler, mat, zou2022compensation} have been designed to incorporate more comprehensive strategies or components. For example, Stadler \etal~\cite{Stadler} employed the objects' active or inactive states to determine the occlusion relationships, subsequently analyzing the velocity of inactive objects to predict their probable positions in the current frame. This approach indirectly models occlusion rather than explicitly estimating it. Conversely, Hibo \etal~\cite{zou2022compensation} proposed a compensation tracker that recovers lost objects using motion compensation. More recently, methods \cite{cao2023observation, yang2024hybrid, wang2025pd, liu2025sparsetrack} attempt to address occlusion in MOT by analyzing the object motion or exploiting depth information. However, these methods still suffer from the cost confusion caused by occlusion, and this problem is more pronounced when objects exhibit similar motions. In this paper, we directly analyze the occlusion severity to mitigate occlusion-induced cost confusion.

\noindent \textbf{Position-and-Feature-Association} combines appearance features \cite{fastreid, ReIDreview, WOS, zhu2022looking, vaswani2017attention} with positional estimations to construct new trackers \cite{deepsort, aharon2022bot, jde, fairmot, ZHOU2020107512, du2023strongsort}, thereby effectively extending the maximum allowable loss time for objects. For instance, Wang \etal \cite{jde} combined FPN \cite{FPN} with a multiscale anchor-based detector to detect objects and extract features. To enhance performance, global and local appearance features \cite{SUN2022190} are extracted using AlphaPose \cite{fang2017rmpe}, which estimates key points (\eg, hands) of objects (\eg, pedestrians) and matches detections with trajectories. Although this approach enhances association robustness, the Position-and-Feature-Association paradigm faces similar challenges regarding positional consistency as the Position-Association paradigm. Consequently, we focus on handling occlusion in the context of Position-Association.
\section{Thinking for Occlusion}\label{sec: 3}
\subsection{Preliminaries}\label{sec: 3.1}
\noindent \textbf{KF} is a linear estimator for dynamical systems. It utilizes the state estimates from the previous time step and the current measurement to predict the object's next state. In tracking, two tasks need to be accomplished: \textit{predict} and \textit{update}. The \textit{update} stage at time $t$ uses the actual observation $\textbf{z}_t$ of a trajectory to obtain the posterior state estimate $\hat{\textbf{x}}_{t|t}$ at time $t$, where the observation corresponds to a detection associated with the trajectory. Mathematically,
\begin{equation}
	\hat{\textbf{x}}_{t|t} = \hat{\textbf{x}}_{t|t-1} + \textbf{\text{K}}_t\left(\textbf{z}_t - \textbf{\text{H}}_t \hat{\textbf{x}}_{t|t-1}\right),
	\label{eq: KF}
\end{equation}
where \textbf{H} denotes the observation model, and \textbf{K} is the Kalman Gain matrix. \textbf{K} is a key process parameters that integrates noise, ensuring the smoothness and stability of estimates. 

\noindent \textbf{IoU} is a critical metric in MOT for assessing the spatial relationship between estimated and detected bounding boxes, describing the position affinity. In this paper, $C_{\mathit{IoU}}(\cdot,\cdot) \in [0,1]$ is used to present the computed IoU value.

\subsection{Influence of Occlusion for Position Association}\label{sec: 3.2}
To begin with, we assume that the practical detected and estimated positions of an object follow $d=P+\Delta_{d}$ and $e=P+\Delta_{e}$, where $P$ denotes the actual position of the object; $\Delta_d$ and $\Delta_e$ indicate the detection and estimation errors. Under partial occlusion, the loss of appearance features increases the detection error, \ie, $\Delta_d\uparrow= \Delta_d + \Delta_{d, occ}$. In a sufficiently short time, the $C_{\mathit{IoU}}\left(d, e\right)$ might change suddenly because of inaccurate detection. Over a long time, due to inaccurate detections occurring frequently, $\Delta_e\uparrow=\Delta_e+\Delta_{e, occ}$ accumulates to become unstable, ultimately resulting in a degradation or fluctuations in $C_{\mathit{IoU}}\left(d, e\right)$. Those phenomena easily cause cost confusion. 

Herein, we assume that object $i$ and $j$ of similar size are spatially close in a moment and object $i$ is occluded by object $j$, \ie, $C_{\mathit{IoU}}(P_i, P_j)$ may approach 1. Due to the instability and accumulation of errors, it might become difficult for the IoU metric to evaluate the spatial consistency between detection and estimation. Moreover, the Hungarian algorithm implements the lowest-position-cost mechanism for detection-estimation allocation, where position cost is indicated as $\mathit{Cost}_{\mathit{IoU}}\left(\cdot,\cdot\right) = 1 - C_{\mathit{IoU}}\left(\cdot,\cdot\right)$. However, due to cumulative errors from inaccurate KF's estimations, $e_i$ and $e_j$, and $d_i$ may change such that $\mathit{Cost}_{\mathit{IoU}}\left(d_j, e_i\right) < \mathit{Cost}_{\mathit{IoU}}\left(d_j, e_j\right)$ or $\mathit{Cost}_{\mathit{IoU}}\left(d_j, e_i\right) < \mathit{Cost}_{\mathit{IoU}}\left(d_i, e_i\right)$, potentially causing false association. This leads to the exchange of identities between trajectories. Our work focuses on observing occlusion to introduce a new cost for objects and optimize KF's parameters. Those approach increases $\mathit{Cost}_{\mathit{IoU}}\left(d_j, e_i\right)$ and decreasing $\mathit{Cost}_{\mathit{IoU}}\left(d_i, e_i\right)$, thereby alleviating cost confusion. 

\section{Methodology}\label{sec:4}
In this section, we introduce \emph{occlusion-aware SORT framework (OA-SORT)} as illustrated in \cref{fig: pipeline}. To address the cost confusion, we integrate the \emph{Occlusion-Aware Module (OAM)} into the position association and the Kalman Filter (KF) updating process. \emph{Occlusion-Aware Offset (OAO)} and \emph{Bias-Aware Momentum (BAM)} assist the association and KF updating, respectively.

\subsection{Occlusion-Aware Module}
\label{sec: OAM}
\subsubsection{Depth Ordering}\label{sec: 4.1.1}
Most two-dimensional cameras implement surveillance in an overhead (non-vertical) view in real-world scenarios. When a camera captures ground-moving objects (\ie, non-flying) on a plane, the relative depth relationship between objects can be estimated from their bounding box bottom edges \cite{quach2024depth, liu2025sparsetrack, wang2025pd}. 
\begin{figure}
	\centering
	\includegraphics[width=\linewidth]{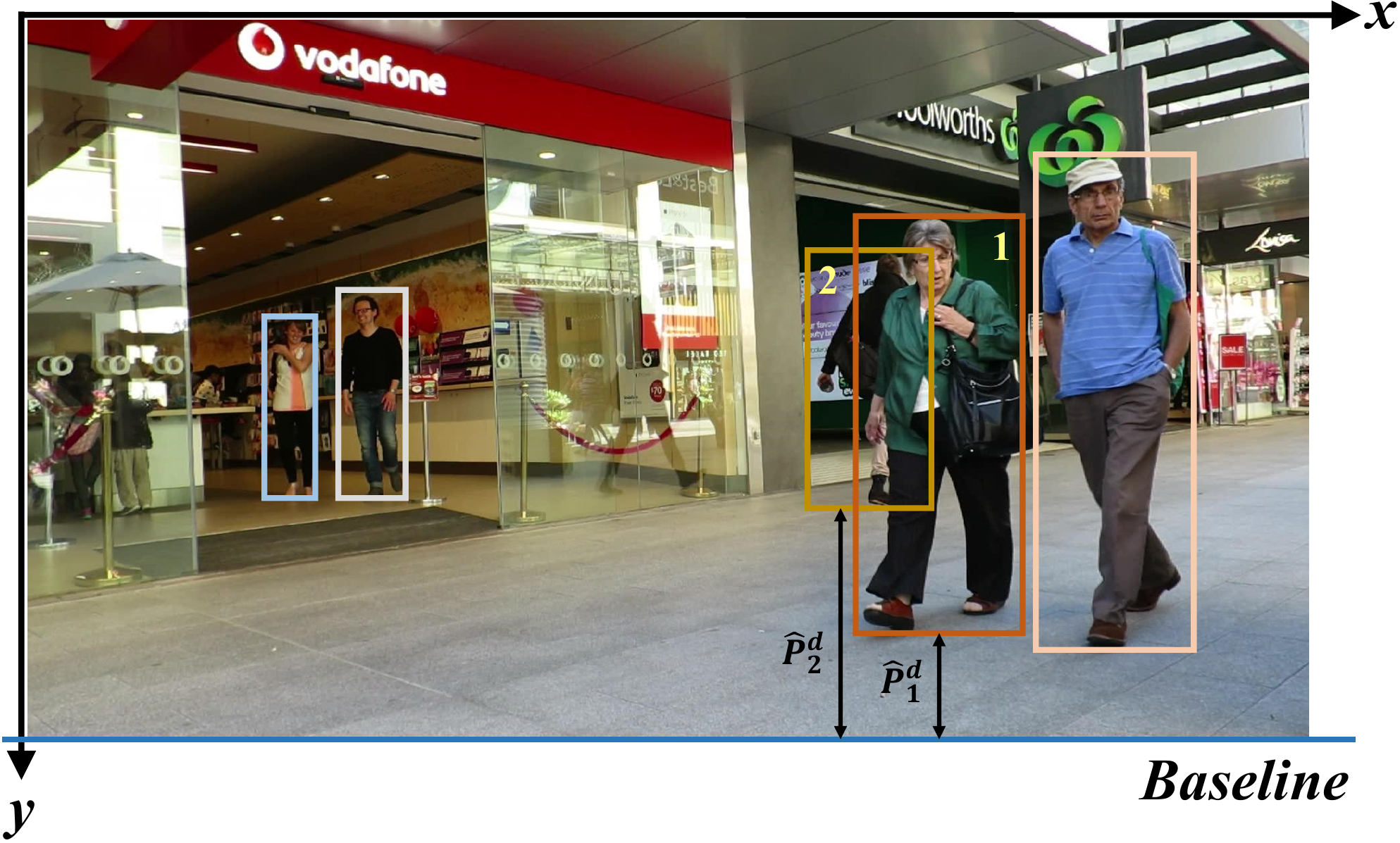}
	\caption{Illustration of the depth relationship. Since the bottom edge of the bounding box is parallel to the \textit{Baseline}, the vertical distance to the \textit{Baseline} is defined as $\hat{P}^d$.}
	\label{fig: obs}
\end{figure}
As illustrated in \cref{fig: obs}, the position of a bounding box's bottom edge provides information about its relative depth with respect to the camera. If $\hat{P}^d_{1} < \hat{P}^d_{2}$, object $\# 1$ is closer to the camera and therefore in front of object $\# 2$. Therefore, for any two detected objects $i$ and $j$,
\begin{equation}
	\left(\hat{P}^d_{i} < \hat{P}^d_{j}\right) \Leftrightarrow \left(\text{Object}\ i \prec \text{Object}\ j\right).
	\label{eq: fbr2}
\end{equation}

This method assist in tracking objects in realistic scenarios and is referred to as \emph{Depth Ordering}. By applying this method, detections can be analyzed to determine the front-and-back relationships between objects. Notably, to avoid fluctuations in the bounding box in actual scenes, a threshold (5) has been set for Depth Ordering to reduce sensitivity, \ie, $\hat{P}^d_{i} + 5 < \hat{P}^d_{j} \rightarrow \text{Object}\ i \prec \text{Object}\ j$.

\subsubsection{Occlusion Coefficient}\label{sec: 4.1.2}
Based on the Depth Ordering, the overlap between detections is calculated to quantify the occlusion severity, expressed as the Occlusion Coefficient ($\mathit{Oc}$). Specifically, let $\mathcal{D}_i$ and $\mathcal{D}_j$ represent the two detections for objects $i$ and $j$, and $\hat{P}^d_{i} > \hat{P}^d_{j}$. The corresponding overlap region is $\mathcal{D}_i \cap \mathcal{D}_j$. Because $\hat{P}^d_{i} > \hat{P}^d_{j}$, object $i$ is located behind object $j$, as defined in Depth Ordering. The $\mathit{Oc} \in [0,1]$ for object $i$ by object $j$ can be expressed as
\begin{equation}
	\mathit{Oc}^j_i = \frac{\textbf{\text{A}}\left(\mathcal{D}_i \cap \mathcal{D}_j\right)}{\textbf{\text{A}}\left(\mathcal{D}_i\right)},
	\label{eq: Oc_1}
\end{equation} 
where $\textbf{\text{A}}(\cdot)$ calculates the pixel count of the region, and the overlap region $\mathcal{D}_i \cap \mathcal{D}_j$ also represents the occlusion region. In practice, an object may be occluded by multiple others. Thus, let $\mathcal{O}_i = \bigcup_{k\in \mathit{Occ}(i)}(\mathcal{D}_i \cap \mathcal{D}_k)$ represent the occluded regions for object $i$, where $\mathit{Occ}(i)$ denotes the index set of objects that occlude object $i$. The global occlusion coefficient $\mathit{Oc_i}$ is interpreted as
\begin{equation}
	\mathit{Oc}_i = \frac{\textbf{\text{A}}\left(\mathcal{O}_i\right)}{\textbf{\text{A}}\left(\mathcal{D}_i\right)}.
	\label{eq: Oc_2}
\end{equation}

The method computes the occlusion coefficient, which serves for bounding boxes generated by the detector or KF.

\subsubsection{Occlusion Coefficient Refinement}
\label{sec: GM}
The occlusion coefficient $\mathit{Oc}$ may overestimate an object's occlusion severity, since its bounding box can include background pixels, particularly near the boundaries. To mitigate background influence, we introduce the Gaussian Map (GM) to refine $\mathit{Oc}$ by adaptively weighting each pixel according to its distance from the bounding box center. For an image frame with $N$ detected objects, the GM value $GM_{x,y} \in [0,1]$ is defined at pixel $(x, y)$ as
\begin{equation}
	\mathit{GM}_{x,y} = \max_{n=1}^{N} \left(e^{-\left(\frac{\left(x-cx_{n}\right)^2}{2\left(\sigma^x_n\right)^2} + \frac{\left(y-cy_{n}\right)^2}{2\left(\sigma^y_n\right)^2}\right)}\right),
	\label{eq: GM}
\end{equation}
where pixel $(x, y)$ is within the detected bounding boxes, $(cx_n, cy_n)$ denotes the centroid of the bounding box of the object $n$, and $(\sigma_n^x, \sigma_n^y)$ are the standard deviations about the detected object category in the horizontal and vertical directions. In this work, $\sigma^x$ and $\sigma^y$ are set proportionally to the bounding box width ($w$) and height ($h$), ensuring smooth decay of the Gaussian kernel from the center to the edges. The refined occlusion coefficient $\hat{\mathit{Oc}}$ for each object $i$ is calculated by
\begin{align}
\hat{\mathit{Oc}}_i 
&= \mathit{Oc}_i \cdot 
\frac{\sum_{\left(x,y\right)\in \mathcal{O}_i} \mathit{GM}_{x,y}}
{\mathbf{A}(\mathcal{O}_i)} \\
&= \frac{\sum_{\left(x,y\right)\in \mathcal{O}_i} \mathit{GM}_{x,y}}
{\textbf{\text{A}}\left(\mathcal{D}_i\right)},
\label{eq:OC}
\end{align}
where pixels closer to the object center dominate the occlusion severity, effectively suppressing the influence of the background. Notably, $\hat{\mathit{Oc}}_i = 0$ when no occlusion occurs. The complete process of obtaining $\hat{\mathit{Oc}}$ constitutes the \textit{Occlusion-Aware Module (OAM)}, corresponding to \textit{OAM w/ GM} in Fig.~\ref{fig: pipeline}.

\subsection{Occlusion-Aware Offset}
\label{sec: OAO}
As discussed above, objects with similar bounding boxes—particularly those in close proximity—exhibit high IoU-based spatial consistency, which can cause position cost confusion. Even when objects are correctly associated with trajectories, their identities may switch due to uncertainty in positional errors. To alleviate this problem, we propose \textit{Occlusion-Aware Offset (OAO)} to integrate the occlusion coefficient into the position cost. Because occlusion often yields inaccurate detections, directly applying OAM to detections is unreliable. This instability arises because the bottom edges of inaccurate detections fluctuate, undermining depth consistency. In contrast, KF not only predicts the position of the trajectory at the next time step using historical motion information but also suppresses the effect of weak noise. Therefore, in this work, OAM is deployed for KF's estimate $\textbf{\text{X}}$ to generate the occlusion coefficient $\hat{\mathit{Oc}}^{\textbf{\text{X}}}$, where $\textbf{\text{X}}$ can be represented using $P+\Delta_e$. By combining $\mathit{C}_{\mathit{IoU}}$ and $\hat{\mathit{Oc}}^{\textbf{\text{X}}}$, the final spatial consistency score $\mathit{S}$ is defined as
\begin{equation}
	\mathit{S} = \tau \cdot \left(1 - \hat{\mathit{Oc}}^{\textbf{\text{X}}}\right) + \left(1 - \tau\right) \cdot \mathit{C}_{\mathit{IoU}}\left(\mathcal{D}, \textbf{\text{X}}\right),
	\label{eq: COST_final}
\end{equation}
where $\tau \in [0, 1]$ represents a coefficient that balances $\mathit{C}_{\mathit{IoU}}\left(\textbf{\text{X}},\mathcal{D}\right)$ and $\hat{\mathit{Oc}}^{\textbf{\text{X}}}$. Thus, $\mathit{Cost}_{\mathit{IoU}}=1-S$ is used to prevent cases where $\mathit{Cost}_{\mathit{IoU}}\left(d_j, e_i\right) < \mathit{Cost}_{\mathit{IoU}}\left(d_j, e_j\right)$. This process, termed \textit{Occlusion-Aware Offset}, operates on KF's predictions and is only triggered during the First-stage association for high-score detections. 

\subsection{Bias-Aware Momentum}
\label{sec: BAM}
Although KF estimates are typically more stable than raw detections, frequent inaccurate detections under occlusion can accumulate errors in the KF, leading to estimation fluctuations. In short time intervals, the KF's estimate is usually more reliable than newly received low-quality detections. Thus, we design \emph{Bias-Aware Momentum (BAM)} for likely inaccurate detections, \ie, those with low scores. BAM incorporates the occlusion coefficient to evaluate the weighting between the estimation and low-score detection, aiming to suppress abnormal variations in the KF's motion parameters. Additionally, to comprehensively account for the spatial relevance, the IoU metric \cite{yang2024hybrid,yu2016unitbox} is incorporated into the BAM calculation to describe the positional relationship between the estimation and the detection. 

It is important to note that BAM is designed to optimize the object's motion estimation. Thus, at time step $t$, OAM is used for a trajectory's latest observation $\textbf{\text{Z}}_{t-1}$ to generate the occlusion coefficient $\hat{\mathit{Oc}}^{\textbf{\text{Z}}_{t-1}}$, while the associated low-score detection is the current observation $\textbf{\textbf{Z}}_t$. $\mathit{BAM}$ is then calculated as 
\begin{equation}
\mathit{BAM} = C_{\mathit{IoU}}\left(\textbf{X}_{t|t-1}, \textbf{\textbf{Z}}_t\right) \cdot \left(1 - \hat{\mathit{Oc}}^{\textbf{\text{Z}}_{t-1}}\right).
\label{eq: BAM}
\end{equation}

Subsequently, during the \textit{update} stage of KF, $\textbf{\textbf{Z}}_t$ is optimized to $\hat{\textbf{Z}}_t$ using $\mathit{BAM}$ as follows:
\begin{equation}	
	\hat{\textbf{\text{Z}}_t} = \mathit{BAM} \cdot \textbf{\textbf{Z}}_t + \left(1-\mathit{BAM}\right) \cdot \textbf{\text{H}}_t \textbf{X}_{t|t-1}.
	\label{eq: bam}
\end{equation}

Finally, the posterior state estimate $\textbf{X}_{t|t}$ at the \textit{update} stage is calculated through
\begin{equation}
	\textbf{X}_{t|t} = \textbf{X}_{t|t-1} + \textbf{\text{K}}_t\left(\hat{\textbf{Z}}_t - \textbf{\text{H}}_t \textbf{X}_{t|t-1}\right).
	\label{eq: z_t}
\end{equation}

Integrating the IoU metrics \cite{yang2024hybrid,yu2016unitbox} ensures that, under the occlusion, the greater the difference between the estimation and detection, the smaller the value of BAM. In other words, as the discrepancy between estimate and detection grows, $\hat{\textbf{Z}}_t$ increasingly depends on the estimation $\textbf{X}_{t|t-1}$. In conclusion, the KF's motion parameters can be dynamically adjusted during the \textit{update} stage to suppress fluctuations.

\subsection{Association approach}\label{sec: app}
OA-SORT adopts Hybrid-SORT as its baseline, \ie, the association includes three stages: (1) the First-stage association links high-score detections with trajectories, (2) the Second-stage handles low-score detections, and (3) the Third-stage reconnects lost trajectories using their latest observation. Overall, our tracking process is shown in \cref{fig: pipeline}. The integration of OAM, OAO, and BAM follows four main steps: 1) After KF estimates for trajectories, OAM utilizes the estimations to calculate the occlusion coefficients; 2) During the process of association, OAO integrates the calculated occlusion coefficients into the spatial consistency metric; 3) For trajectories associated with low-score detection, BAM uses the occlusion coefficients of trajectories' latest observations to optimize the KF's motion parameters; 4) Before the end of the current frame, OAM utilizes the trajectories' latest observation to calculate the occlusion coefficient for subsequent BAM. In practice, the tracking framework can be integrated into other trackers~\cite{zhang2022bytetrack, cao2023observation, liu2025sparsetrack, wang2025pd, sort, shim2025focusing}. Notably, both OAO and BAM modules can operate independently.
\section{Experiment}
\label{sec: exp}
\subsection{Experimental Setup}
\noindent \textbf{Datasets}. We evaluate OA-SORT on three widely used benchmarks: DanceTrack~\cite{DanceTrack}, SportsMOT~\cite{cui2023sportsmot}, and MOT17~\cite{MOTChallenge}. DanceTrack is a challenging benchmark for MOT in stage performance scenarios. It includes diverse nonlinear motions and frequent partial occlusions, whereas SportsMOT features variable-speed motion and dynamic camera movement. MOT17 comprises pedestrian street scenes with frequent and prolonged occlusions.

\noindent \textbf{Metrics}. The metrics in the experiments are defined in CLEAR \cite{2008Evaluating}, HOTA \cite{luiten2020IJCV}, and PMDS \cite{PMDS}. Here, $\uparrow$ and $\downarrow$ respectively denote that higher or lower values correspond to better performance: \textbf{HOTA (\%) $\uparrow$} is a higher-order tracking accuracy metric, which represents the performance of detection, association, and localization; \textbf{MOTA (\%) $\uparrow$} is a comprehensive evaluation indicator for MOT in combination with missed detection, false detections, and ID switch; \textbf{IDF1 (\%) $\uparrow$} measures the accuracy of assigning identities; \textbf{AssA (\%) $\uparrow$} measures association accuracy, which is simply the average alignment between matched trajectories.

\noindent \textbf{Implementation Details}. The hyperparameters of OA-SORT are the same as those in Hybrid-SORT~\cite{yang2024hybrid}. In OAM, $\sigma^x $ is $w/3\sqrt{2}$ and $\sigma^y$ is $h/3$ for DanceTrack; $\sigma^x$ is $w/4$ and $\sigma^y$ is $h/3$ for SportsMOT; $\sigma^x$ is $w/2$ and $\sigma^y$ is $h/2$ for MOT17. In OAO, $\tau$ is 0.15 for DanceTrack, 0.2 for SportsMOT, and 0.1 for MOT17. $\tau$ is empirically tuned between 0.1 and 0.2 to balance precision and robustness across motion patterns. Herein, $\tau$ for SportsMOT and MOT17 is empirically set based on average IoU between adjacent frames \cite{cui2023sportsmot}. All experiments are conducted on a single NVIDIA V100 GPU paired with an Intel Xeon (R) 6130 CPU (2.10 GHz).

\subsection{Benchmark Results}
\noindent \textbf{DanceTrack.} As shown in \cref{tab: dancetrack}, OA-SORT achieves consistent gains over the baseline Hybrid-SORT~\cite{yang2024hybrid} (+0.9 HOTA, +1.1 AssA, +0.1 MOTA, +1.2 IDF1). Improvements in AssA and IDF1 indicate that the occlusion-aware framework effectively mitigates cost confusion and refines KF estimation. Notably, Hybrid-SORT already exploits motion direction cues to mitigate occlusion effects. Additionally, integrating occlusion-awareness into other trackers—ByteTrack (OA-Byte), OC-SORT (OA-OC), SparseTrack (OA-Sparse), and PD-SORT (OA-PD)—consistently improves their performance. Although PD-SORT \cite{wang2025pd} and SparseTrack \cite{liu2025sparsetrack} leverage pseudo-depth to design spatial consistency index metric (DVIoU) and association strategy (DCM), occlusion-aware framework integration further improves performance. Overall, the results demonstrate that modeling and leveraging occlusion states effectively handle nonlinear and interactive motion scenarios.

\begin{table}[t!]
\small
\centering
\begin{tabular}{lcccc}
\toprule
Method                                                               & HOTA             & AssA            & MOTA              & IDF1             \\
\midrule
CenterTrack \cite{Centertrack}               & 41.8             &  22.6           & 86.8              & 35.7             \\
FairMOT \cite{fairmot}                       & 39.7	            & 23.8            & 82.2              & 40.8             \\
QDTrack \cite{QDTrackX}                      & 45.7             & 29.2            & 83.0              & 44.8             \\
DST-Tracker \cite{cao2022track}              & 51.9	            & 34.6            & 84.9              & 51.0             \\
FineTrack \cite{ren2023focus}                & 52.7             & 38.5            & 89.9              & 59.8             \\
DiffusionTrack \cite{luo2024diffusiontrack}  &52.4              &33.5             &89.3               &47.5              \\
MOTIP          \cite{gao2025multiple}        & 67.5             & 57.6            & 90.3              &72.2              \\
\midrule
\multicolumn{5}{l}{\textit{Same Detection:}}                                                                                                     \\
\rowcolor{blue2}		SORT \cite{sort}                             & 47.9             & 31.2            & \underline{91.8}  & 50.8             \\ 
\rowcolor{blue2}		StrongSORT++ \cite{du2023strongsort}         & 55.6             & 38.6            & 91.1              & 55.2             \\
\rowcolor{blue2}		C-BIoU \cite{yang2023hard}                   & 60.6             & 45.4            & 91.6              & 61.6             \\
\rowcolor{blue2}		MotionTrack2024 \cite{xiao2024motiontrack}   & 58.2             & 41.7            & 91.3              & 58.6             \\
\rowcolor{blue2}		ETTrack \cite{han2025ettrack}                & 56.4             & 39.1            & \textbf{92.2}     & 57.5             \\
\rowcolor{blue2}		C-TWiX \cite{miah2025learning}               & 62.1             & 47.2            & 91.4              & \underline{63.6} \\
\rowcolor{blue2}		\multicolumn{5}{l}{\textbf{\textit{Integration:}}}                                                                       \\
\rowcolor{blue2}	    ByteTrack \cite{zhang2022bytetrack}          & 47.1             & 31.5            & 88.2              & 51.9             \\ 
\rowcolor{blue2}		OA-Byte (Ours)                               & 49.0             & 33.7            & 89.6              & 55.9             \\
\rowcolor{blue2}		OC-SORT \cite{cao2023observation}            & 54.6             & 40.2            & 89.6              & 54.6             \\
\rowcolor{blue2}		OA-OC (Ours)                                 & 56.5             & 39.6            & 91.2              & 57.6             \\
\rowcolor{blue2}		SparseTrack \cite{liu2025sparsetrack}        & 55.5             & 39.1            & 91.3              & 58.3             \\
\rowcolor{blue2}		OA-Sparse (Ours)                             & 57.8             & 41.8            & 91.5              & 60.2             \\
\rowcolor{blue2}		PD-SORT \cite{wang2025pd}                    & 58.2             & 42.1            & 89.6              & 57.5             \\
\rowcolor{blue2}		OA-PD (Ours)                                 & 60.4             & 44.9            & 91.4              & 60.8             \\
\rowcolor{blue2}		Hybrid-SORT \cite{yang2024hybrid}            & \underline{62.2} & \underline{47.4}& 91.6              & 63.0             \\
\rowcolor{blue2}		OA-SORT (Ours)                               & \textbf{63.1}    &  \textbf{48.5}  & 91.7              & \textbf{64.2}    \\
\bottomrule
\end{tabular}
\caption{Results on DanceTrack test set. The same detection is provided by OC-SORT.}
\label{tab: dancetrack}
\end{table}

\noindent \textbf{SportsMOT.} \cref{tab: SportsMOT} presents results on SportsMOT. The results show that the occlusion-aware framework remains effective under variable-speed motion and camera-view changes. This effectiveness arises because occlusion evaluation is based on relative positions between objects. Even after short-term camera view changes, relative position can be maintained. Notably, OA-SORT achieves these results without any camera-motion compensation. Additionally, as noted in \cite{cui2023sportsmot}, SportsMOT imposes stricter requirements on IoU and Kalman filter accuracy than DanceTrack. The results further confirm the effectiveness and generalizability of the proposed framework. It effectively enhances tracking robustness under occlusion scenarios involving variable-speed motion and high-intensity camera movement.

\begin{table}[t!]
\small
\centering
\begin{tabular}{lcccc}
\toprule
Method                                                          & HOTA$\uparrow$   & AssA$\uparrow$   & MOTA$\uparrow$    & IDF1$\uparrow$ \\
\hline
FairMOT \cite{fairmot}                                          & 49.3             & 34.7             & 86.4              & 53.5             \\
QDTrack \cite{QDTrackX}                                         & 60.4             & 47.2             & 90.1              & 62.3             \\
TransTrack \cite{sun2020transtrack}                             & 68.9             & 57.5             & 92.6              & 71.5             \\
MeMOTR \cite{gaoMeMOTRLongTermMemoryAugmented2023}              & 70.0             & 59.1             & 91.5              &71.4              \\
MOTIP \cite{gao2025multiple}                                    & 71.9             & 62.0             & 92.9              &75.0              \\
\midrule
\multicolumn{5}{l}{\textit{Same Detection:}}                                                                                                 \\
\rowcolor{blue2}		ByteTrack \cite{zhang2022bytetrack}     & 62.8             & 51.2             & 94.1              & 69.8             \\
\rowcolor{blue2}		OC-SORT \cite{cao2023observation}       & 71.9             & 59.8             & 94.5              & 72.2             \\
\rowcolor{blue2}		DiffMOT \cite{lv2024diffmot}            & 72.1             & 60.5             & 94.5              & 72.8             \\
\rowcolor{blue2}		MambaTrack \cite{xiao2024mambatrack}    & 72.6             & 60.3             & \textbf{95.3}     & 72.8             \\
\rowcolor{blue2}		Hybrid-SORT \cite{yang2024hybrid}       & \underline{73.0} & \underline{61.6} &  94.3             & \underline{73.3} \\  
\rowcolor{blue2}		OA-SORT (Ours)                          & \textbf{73.4}    &   \textbf{62.3}  &  \underline{94.4} & \textbf{74.1}    \\
\hline
\multicolumn{5}{l}{*\textit{Same Detection:}}                                                                               \\
\rowcolor{blue3}		ByteTrack \cite{zhang2022bytetrack}     & 64.1             & 52.3             & 95.9             & 71.4              \\
\rowcolor{blue3}		MixSort-Byte \cite{cui2023sportsmot}    & 65.7             & 54.8             & 96.2             & 74.1              \\
\rowcolor{blue3}		OC-SORT \cite{cao2023observation}       & 73.7             & 61.5             & \textbf{96.5}    & 74.0              \\
\rowcolor{blue3}		MixSort-OC \cite{cui2023sportsmot}      & 74.1             & 62.0             & \textbf{96.5}    & 74.4              \\
\rowcolor{blue3}		Hybrid-SORT \cite{yang2024hybrid}       & \underline{74.8} & \underline{63.2} & 96.2             & \underline{75.1}  \\  
\rowcolor{blue3}		OA-SORT (Ours)                          & \textbf{75.2}    & \textbf{63.8}    & \underline{96.3} & \textbf{75.8}     \\
\bottomrule
\end{tabular}
\caption{Results on SportsMOT. The same detections are provided by SportsMOT, where * indicates that the detector is trained using the SportsMOT train and validation sets.}
\label{tab: SportsMOT}
\end{table}

\noindent \textbf{MOT17.} \cref{tab: mot17} presents the results on MOT17, which represents a more general and typical scenario of linear-motion patterns. OA-SORT still surpasses the baseline, Hybrid-SORT (+0.6 HOTA and +0.7 IDF1), even outperforms Hybrid-SORT-REID by +0.5 AssA and +0.3 IDF1. These results demonstrate that occlusion-awareness is highly versatile across different motion patterns. Moreover, the occlusion-awareness brings benefits for BOT-SORT without ReID (OA-BOT). 
\begin{table}[t]
\small
\centering
\begin{tabular}{lcccc}
\toprule
Method                                                         & HOTA           & AssA           & MOTA           & IDF1             \\
\midrule  
CenterTrack~\cite{Centertrack}                                 & 52.5           & 51.0           & 67.8           & 64.7             \\
FairMOT~\cite{fairmot}                                         & 59.3           & 58.0           & 73.7           & 72.3             \\
DST-Tracker~\cite{cao2022track}                                & 60.1           & 62.1           & 75.2           & 72.3             \\
UTM~\cite{you2023utm}                                          & 64.0           & 62.5           & 81.8           & 78.7             \\
DiffusionTrack~\cite{luo2024diffusiontrack}                    & 60.8           & 58.8           & 77.9           &73.8              \\
MOTIP~\cite{gao2025multiple}                                   & 59.2           & 56.9           & 75.5           &71.2              \\
\midrule
\multicolumn{5}{l}{\textit{Same Detection:}}                                                                                         \\
\rowcolor{blue2} ByteTrack~\cite{zhang2022bytetrack}           & 63.1           & 62.0           & 80.3           & 77.3             \\
\rowcolor{blue2} OC-SORT~\cite{cao2023observation}             & 63.2           & 63.2           & 78.0           & 77.5             \\
\rowcolor{blue2} StrongSORT++~\cite{du2023strongsort}          & 64.4           & 64.4           & 79.6           & 79.5             \\
\rowcolor{blue2} PD-SORT~\cite{wang2025pd}                     & 63.9           & 64.1           & 79.3           & 79.2             \\
\rowcolor{blue2} Hybrid-SORT-REID~\cite{yang2024hybrid}        & 64.0           & 63.5           & 79.9           & 78.7             \\
\rowcolor{blue2} BOT-SORT-REID~\cite{aharon2022bot}            &\underline{65.0}& \textbf{65.5}  & 80.5           & \textbf{80.2}    \\
\rowcolor{blue2} SMILEtrack~\cite{wang2024smiletrack}          &\underline{65.0}& -              &\underline{80.7}& \underline{80.1} \\
\rowcolor{blue2} \multicolumn{5}{l}{\textbf{\textit{Integration:}}}                                                                  \\
\rowcolor{blue2} BOT-SORT~\cite{aharon2022bot}                 & 64.6           & -              & 80.6           & 79.5             \\  
\rowcolor{blue2} OA-BOT (Ours)                                 & \textbf{65.1}  &\underline{64.8}& \textbf{80.9}  & 79.4             \\  
\rowcolor{blue2} Hybrid-SORT~\cite{yang2024hybrid}             & 63.6           & 63.2           & 80.6           & 78.4             \\  
\rowcolor{blue2} OA-SORT (Ours)                                & 64.2           & 64.0           &  79.6          & 79.1             \\
\bottomrule
\end{tabular}
\caption{Results on MOT17-test with the private detections. The same detection is provided by ByteTrack.}
\label{tab: mot17}
\end{table}

Overall, integrating occlusion-awareness into the tracking framework achieves better performance. These results also indicate that the occlusion-aware framework enhances tracking robustness under different occlusion scenarios, involving variable-speed and camera motion, validating the rationality. Even on MOT17, which includes numerous false detections and long-term missed detections, there have been improvements. Additionally, we evaluate Hybrid-SORT on MOT20 \cite{dendorfer2020mot20}, achieving an additional +0.4 IDF1 improvement. Notably, optimizing the number of missed and false detections is not the primary focus of this work. 
\subsection{Ablation Study} \label{Sec:5.2}
Detections and ReID features, provided by Hybrid-SORT~\cite{yang2024hybrid,fastreid,ge2021yolox}, are utilized in this section.

\noindent \textbf{Component Ablation.} The results for OAO, BAM, and GM are reported in \cref{tab: componentDance}. These results demonstrate that introducing occlusion-awareness significantly improves tracking association and accuracy. Along with integrating OAO, the tracker's association is enhanced, with improvements of +0.5 HOTA. Secondly, after optimizing KF's estimate through BAM, the HOTA is significantly improved by +1.1, which has only a minor impact on inference speed (average +3.81ms). Notably, integrating GM brings improvements of +2.1 HOTA. Although GM increases the average tracking time to 24.56 ms per frame, the system still satisfies real-time tracking requirements. Additionally, with the use of occlusion-awareness, Hybrid-SORT-REID also gets benefits. Notably, given $X$ trajectories with $Z$ interactions and $N$ detected objects with $M$ interactions, the computational complexity is approximately: OAM: $\mathcal{O}(X^2 + Z * \hat{A})$ or $\mathcal{O}(N^2 + M * \hat{A})$, OAO: $\mathcal{O}(Z * N)$, and BAM: $\mathcal{O}(M)$, where $\hat{A}$ denotes the average pixel area of the detected bounding boxes. 

\begin{table}[t!]
 \small
\setlength{\tabcolsep}{1.2mm}
\centering
\begin{tabular}{cccc|cccc}
\toprule
OAO             &      BAM     & GM          & ReID       &   HOTA         &  AssA & IDF1    & T (ms)   \\
\midrule
				&              &             &            & 59.4           & 44.9  & 60.7    & 9.57     \\
\checkmark      &              &             &            & 59.9           & 45.6  & 60.9    & +1.96    \\
\checkmark      &              & \checkmark  &            & 60.6           & 46.7  & 62.0    & +9.25    \\
\checkmark      & \checkmark   &             &            & 60.5           & 46.5  & 62.2    & +3.81    \\
\checkmark      & \checkmark   &  \checkmark &            & 61.5           & 48.0   & 63.7   & +14.99   \\
\midrule
				&              &             & \checkmark & 63.1           & 50.4   & 65.1   & 60.95    \\
\checkmark      & \checkmark   &  \checkmark & \checkmark & 63.4           & 50.8   & 65.3   & +15.32   \\
\bottomrule 
\end{tabular}
\caption{Components ablation on DanceTrack-val; T (ms) indicates average tracking runtime per frame. }
\label{tab: componentDance}
\end{table}
\noindent \textbf{$\sigma^x$ and $\sigma^y$ in GM.} Smaller $\sigma^x$and $\sigma^y$values concentrate the weights more heavily at the center of the bounding box, reducing edge influence. For pedestrian tracking, the optimal parameters may vary according to the motion pattern. For example, the Dancetrack dataset primarily consists of dance scenes, and the change in hand and leg positions introduces a significant amount of background; therefore, the horizontal and vertical weights should be smaller. Experimental results confirm our analysis that $\sigma^x = w/3\sqrt{2}$ and $\sigma^y=h/3$ for DanceTrack can achieve good performance.

\begin{figure}[t]
\centering
\includegraphics[width=\linewidth]{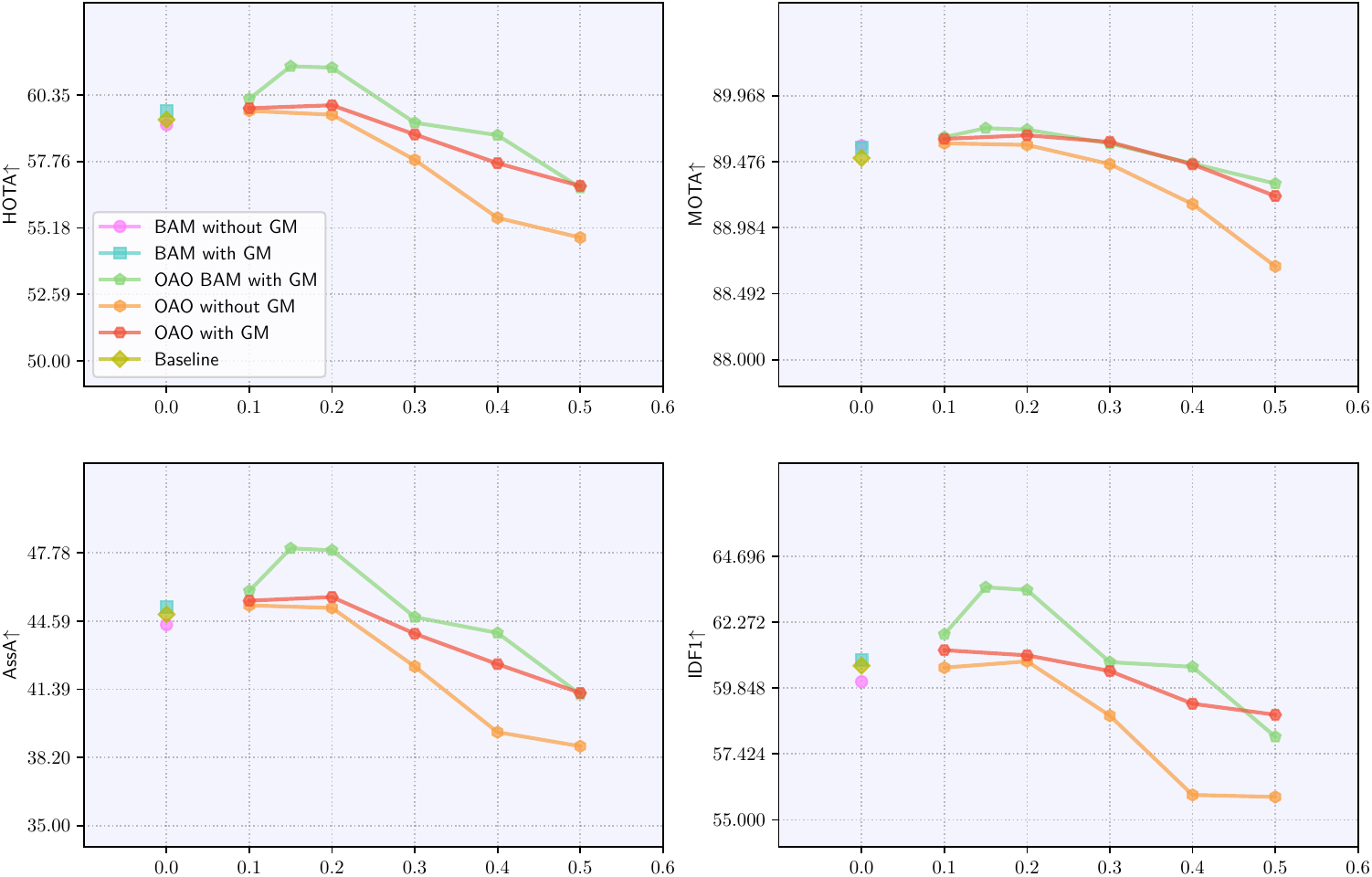}
\caption{The results under GM and different values of $\tau$.}
\label{fig: Various}
\end{figure}

\noindent \textbf{GM and $\tau$.} A larger $\tau$ as defined in \cref{eq: COST_final} mitigates the position cost confusion due to inaccurate detection caused by occlusion, but is also likely to destroy the spatial consistency expression. Thus, we show \cref{fig: Various} that the tracking performance of OAO and BAM components under different GM and values of $\tau \in [0.1, 0.5]$, which analyzes the influence of different $\tau$ for different configurations. Overall, our results agree with our analysis that increasing $\tau$ from $\tau = 0.1$ can yield benefits using the tracker with OAO, BAM, and GM on DanceTrack. However, increasing $\tau$ beyond a certain value degrades performance due to compromised spatial consistency. Secondly, GM effectively enhances the performance of both OAO and BAM components, regardless of the $\tau$. 

\noindent \textbf{Bias-Aware Momentum.} To mitigate the impact of inaccurate detections, BAM (\cref{sec: BAM}) combines the spatial-consistency metric~\cite{yang2024hybrid,yu2016unitbox} and occlusion information. There are alternative configurations, such as using a constant value instead of the spatial consistency metric~\cite{yang2024hybrid,yu2016unitbox}. The results in \cref{tab: BAM} indicate that incorporating the spatial consistency metric~\cite{yang2024hybrid,yu2016unitbox} for adaptive momentum adjustment achieves better performance. To reduce the extra computation, the spatial consistency metric can follow the tracker. Herein, following Hybrid-SORT~\cite{yang2024hybrid}, OA-SORT utilizes HMIoU.
\begin{table}[t!]
\small
\centering
\begin{tabular}{c|ccc}
\toprule
                             & HOTA              & AssA              & IDF1              \\
\midrule
0.1                          & 60.69             & 46.78             & 62.47             \\
0.3                          & 60.99             & 47.16             & 62.99             \\
0.4                          & \underline{61.26} & \underline{47.61} & \underline{63.00} \\
0.45                         & 61.04             & 47.27             & 62.65             \\
0.5                          & 61.02             & 47.24             & 62.77             \\
IoU~\cite{yu2016unitbox}     & \textbf{61.49}    & \textbf{48.03}    & \textbf{63.65}    \\
HMIoU~\cite{yang2024hybrid}  & \textbf{61.49}    & \textbf{48.03}    & \textbf{63.65}    \\
\bottomrule
\end{tabular}
\caption{Results on DanceTrack-val with different BAM.}
\label{tab: BAM}
\end{table}

\noindent \textbf{Application on Other Trackers.} We incorporated OAO and BAM with GM into different trackers, and the results are shown in \cref{tab: Application}. Consequently, OAO and BAM consistently enhance each tracker's performance, demonstrating their adaptability and reusability. Notably, the four trackers have different association methods and strategies. The TrackTrack~\cite{shim2025focusing} employs additional detections, ReID feature~\cite{fastreid}, novel allocation method differing from the Hungarian algorithm, and the NSA Kalman Filter~\cite{GIAOTracker} that is an extended Kalman filter. Its association process for confirmed trajectory is between OAO and BAM.
\begin{table}[t!]
\small
\centering
\begin{tabular}{c|ccc|cc}
\toprule
Tracker                                             &  Byte      &     OAO       &       BAM       & HOTA        &  IDF1           \\
\midrule
\multirow{4}{*}{SORT \cite{sort}}                   &            &               &                 & 48.4        &  49.6           \\
                                                    &  \checkmark&               &                 & 48.3        &  50.0           \\
                                                    &  \checkmark&  \checkmark   &                 & 50.4        &  53.0           \\
                                                    &  \checkmark&  \checkmark   & \checkmark      & 50.4        &  53.3           \\
\midrule
\multirow{4}{*}{Bytetrack \cite{zhang2022bytetrack}}&     -      &      -        &    -            &    -        &  -              \\
                                                    &  \checkmark&               &                 &      47.1   &  52.7           \\
                                                    &  \checkmark&  \checkmark   &                 & 48.8        &  55.1           \\
                                                    &  \checkmark&  \checkmark   & \checkmark      & 49.3        &  55.7           \\
\midrule
\multirow{4}{*}{OC-SORT \cite{cao2023observation}}  &            &               &                 &   52.3      &  52.0           \\
                                                    &  \checkmark&               &                 &  52.4       &  51.5           \\
                                                    &  \checkmark&  \checkmark   &                 &  53.6       &  53.3           \\
                                                    &  \checkmark &  \checkmark  & \checkmark      &  53.8       &  53.9           \\
\midrule
\multirow{4}{*}{TrackTrack \cite{shim2025focusing}} &      -     &     -         &      -          &   -         &  -              \\
                                                    &  \checkmark&               &                 &  59.3       &  61.1           \\
                                                    &  \checkmark&  \checkmark   &                 &  59.7       &  61.7           \\
                                                    &  \checkmark &  \checkmark  & \checkmark      &  59.9       &  62.0           \\
\bottomrule
\end{tabular}
\caption{Results on DanceTrack-val. To highlight BAM, the low-score detection association (Byte) is integrated.}
\label{tab: Application}
\end{table}

\subsection{Limitations}\label{sec:6}
Overall, the occlusion-aware framework has stronger robustness in occluded scenes. However, when the lower parts of objects are occluded or when objects are airborne (\eg, jumping), the framework's association performance decreases compared to the baseline. As shown in \cref{fig:occlusion}, sequence \#0026 serves as a representative example. In this case, the bottom-edge-based method struggles to accurately capture the depth relationships, thereby affecting the performance of occlusion-aware framework. Secondly, the occlusion state and its temporal variations are continuous over time. Although this paper reveals that observing and utilizing occlusion state can enhance association robustness, the lack of long-term occlusion modeling leads to instability in association. This issue will be addressed in future research.
\section{Conclusion}\label{sec:7}
In this paper, we explore the impact of occlusion on position association. Inaccurate detection and cost confusion caused by occlusion seriously affect tracking accuracy. However, existing trackers have not modeled the occlusion status of bounding box detections. To address this, the proposed occlusion-aware framework integrates occlusion status into the position association and the update phase of KF. Notably, GM is introduced to construct OAM, enabling more accurate evaluation of occlusion severity. Comprehensive experiments conduct on multiple datasets fully validate the importance of integrating occlusion states. In terms of the ablation study, the role, practicality, and reusability of GM, OAO, and BAM are thoroughly demonstrated. In summary, the three plug-and-play, training-free components are easily integrated into existing trackers. In future work, we will explore more robust occlusion state estimation methods and develop low-complexity approaches for estimating occlusion severity.
\section*{Acknowledgments}
This work was supported in part by National Natural Science Foundation of China (Grant 62302324); in part by Sichuan Province Science and Technology Support Program (Grant 2024NSFSC0500); in part by the Fundamental Research Funds for the Central Universities (Grant YJ202420); in part by Sichuan University young teachers science and technology innovation ability improvement project (Grant 2024SCUQJTX028).

{
    \small
    \bibliographystyle{ieeenat_fullname}
    \bibliography{reference}
}
\clearpage
\setcounter{page}{1}
\maketitlesupplementary

\section{Reliability Analysis and Pseudo code}
\label{sec:rationale}
\subsection{Occlusion-Aware Module}
\label{OAM}
\subsubsection{Reliability Analysis}
Occlusion in 2D image space follows two primary spatial cues:
\begin{itemize}
\item Depth ordering: In pedestrian or vehicle scenes, 2D overlap typically indicates physical proximity, and bounding boxes with significantly lower bottom-edge y-coordinates (\ie, larger $b$, \cref{fig: obs}) are typically nearer to the camera and thus more likely to occlude others.
\item Intersection area: For objects with intersecting bounding boxes (non-zero IoU), the overlapping region correlates with the extent of physical occlusion.
\end{itemize}

OAM leverages both cues and constructs a logical occlusion relationship matrix:
\begin{equation}
C_\mathit{IoU}\left(\mathcal{D}_i,\mathcal{D}_j\right) > 0 \land b_j - b_i > \mathit{thre}_{occ}
\end{equation}
where $C_\mathit{IoU}(\mathcal{D}_i,\mathcal{D}_j)$ ensures that only \textbf{geometrically consistent occlusion hypotheses} are considered; $\mathit{thre}_{occ}$ aims to reduce \textbf{false depth ordering brought by fluctuations and strong competition at the bottom edge}.

In terms of GM (\cref{eq: GM}), for each bounding box $i$, $\sigma^x_i$ and $\sigma^y_i$ are defined by:
\begin{equation}
	\sigma^x_i = \frac{w_i}{k_x},\quad \sigma^y_i = \frac{h_i}{k_y}.
\end{equation}

The GM aims to reflect the relative importance of pixels affected by occlusion. All in all, OAM applies a soft probabilistic prior that reflects the confidence that multiple trajectories should be associated. This design complements hard IoU logic from the trajectory view, emphasizing the association of unoccluded trajectories.

\subsubsection{Pseudocode}
$\mathcal{D}=[l, t, r, b]$ indicates detected bounding boxes, where $l$, $t$, $r$, $b$ denote the left, top, right, and bottom of the bounding box. Herein, we assume that $N$ bounding boxes ($\mathcal{D}\in \mathbb{R^*}^{N\times4}$), need to be associated with high-score detections, where the width and height of the image are W and H. The Pseudo-code about OAM is shown as follows, where $\leftarrow$ indicates assignment.
Algorithm \ref{alg:oam} implements three key components:
\begin{itemize}
	\item Lines 5-9: \cref{eq: GM};
	\item Lines 10-35: \cref{eq: fbr2} and \cref{eq: Oc_2};
	\item To improve computational efficiency, lines 2-3 and line 6 are used for data filtering.
\end{itemize}

Moreover, it is worth noting that to avoid fluctuations in estimation or detection, we set a threshold ($thre_{occ}$) for comparing the bottom edges (line 12 of the algorithm). When the difference value exceeds $thre_{occ}$, it will be considered an occlusion. The $thre_{occ}$ for the experiment in our work is 5. This can also alleviate the corresponding issues caused by incorrect bottom-edge depth ordering. Moreover, Gaussian scaling factors $k_x$ and $k_y$ are used to adapt to the proportion of background pixels within each detected bounding box. Their values are discussed in the \cref{sec:gm}.

\begin{algorithm}[t]
	\small
	\caption{Occlusion-Aware Module ($\hat{\mathit{Oc}} \leftarrow OAM(\cdot)$)}
	\label{alg:oam}
	\textbf{Input}: Detected bounding boxes $\mathcal{D} \in \mathbb{R}^{N\times4}$\\
	\textbf{Parameter}: Image dimensions $(W, H)$, Gaussian scaling factors $k_x, k_y$, occlusion trigger threshold $thre_{occ}$ \\
	\textbf{Output}: Refined occlusion coefficients $\hat{\mathit{Oc}} \in \mathbb{R}^{N}$
	
	\begin{algorithmic}[1]
		\STATE Initialize $\hat{\mathit{Oc}} \leftarrow \mathbf{0}^{N}$ \COMMENT{Initialize occlusion coefficients}
		\STATE $IoU_{\mathcal{D}} \in \mathbb{R}^{N\times N} \leftarrow C_{IoU}(\mathcal{D}, \mathcal{D})$ \COMMENT{Compute IoU matrix}
		\STATE $\text{diag}(IoU_{\mathcal{D}}) \leftarrow 0$ \COMMENT{Ignore self-relation}
		\\ \# --------------\textbf{\textit{Spatial consistency determination}}-------------
		\IF{$\max(IoU_{\mathcal{D}}) > 0$}
		\STATE $GM \in \mathbb{R}^{H \times W} \leftarrow \mathbf{0}^{H \times W}$ \COMMENT{Initialize Gauss-Map}
		\FOR{each $\mathcal{D}_i$ in $\{\mathcal{D}_i | \max(IoU_{\mathcal{D}}[i]) > 0\}$} 
		\STATE $\sigma^x_i \leftarrow w_i / k_x,\ \sigma^y_i \leftarrow h_i / k_y$ \COMMENT{Adaptive sigmas}
		\STATE $GM[t_i:b_i, l_i:r_i] \leftarrow \max(GM[t_i:b_i, l_i:r_i], \text{drawGaussian}(\mathcal{D}_i, \sigma^x_i, \sigma^y_i))$ 
		\ENDFOR
		\STATE $bottoms \leftarrow \mathcal{D}[:,3]$ \COMMENT{Bottom y-coordinates ($b$)}
		\STATE $areas \leftarrow (\mathcal{D}[:,2] - \mathcal{D}[:,0]) \times (\mathcal{D}[:,3] - \mathcal{D}[:,1])$ \COMMENT{Calculate Bounding-box areas}
		\STATE $validMask \leftarrow (bottoms[:, None] - bottoms[None, :] \leq -thre_{occ}) \land (IoU_{\mathcal{D}} > 0)$
		\COMMENT{The occlusion relationship matrix.}
		\STATE $all\_L, all\_T, all\_R, all\_B \leftarrow \mathcal{D}$ \COMMENT{Vectorized coordinates}
		\\ \# ---------------\textbf{\textit{Obtaining occlusion coefficient}}--------------
		\FOR{each $\mathcal{D}_i$ and $i$ in $\mathcal{D}$}
		\STATE $js \leftarrow \{ j \mid validMask[i,j] = \text{TRUE} \}$ \COMMENT{Select objects that occludes object $i$.}
		\IF{$js = \emptyset$} 
		\STATE  \textbf{continue}
		\ENDIF
		\STATE $l_i, t_i, r_i, b_i \leftarrow \mathcal{D}_i$
		\STATE $localGM \in \mathbb{R}^{h_i \times w_i} \leftarrow GM[t_i:b_i, l_i:r_i]$ \COMMENT{Crop local GM from $GM$: Gaussian heatmap about $\mathcal{D}_i$}
		
		\STATE $T\in \mathbb{R}^{len(js)} \leftarrow \max(t_i, all\_T[js])$ \COMMENT{$len(js)$ is array length of $js$}
		\STATE $B\in \mathbb{R}^{len(js)} \leftarrow \min(b_i, all\_B[js])$
		\STATE $L\in \mathbb{R}^{len(js)} \leftarrow \max(l_i, all\_L[js])$
		\STATE $R\in \mathbb{R}^{len(js)} \leftarrow \min(r_i, all\_R[js])$
		
		\STATE $tClip\in \mathbb{R}^{len(js)} \leftarrow \max(0, T - t_i)$   \COMMENT{Coordinate transformation}
		\STATE $bClip\in \mathbb{R}^{len(js)} \leftarrow \min(h_i, B - b_i)$
		\STATE $lClip\in \mathbb{R}^{len(js)} \leftarrow \max(0, L - l_i)$
		\STATE $rClip\in \mathbb{R}^{len(js)} \leftarrow \min(w_i, R - r_i)$
		
		\STATE $occlusionMap \leftarrow \mathbf{0}^{h_i \times w_i}$ \COMMENT{Boolean occlusion map: occlusion region}
		\FOR{each $j$ in $js$}
		\STATE $occlusionMap[tClip[j]:bClip[j], lClip[j]:rClip[j]] \leftarrow 1$ \COMMENT{Obtain the overlapping region}
		\ENDFOR
		\STATE $\hat{\mathit{Oc}}_i \leftarrow \frac{\sum\left(localGM \odot occlusionMap\right)}{areas_i}$  \COMMENT{$\odot$ indicates Hadamard (element-wise) product.}
		\ENDFOR
		\ENDIF
		\STATE \textbf{return} $\hat{\mathit{Oc}}$
	\end{algorithmic}
\end{algorithm}

\subsection{Occlusion-Aware Offset}
\label{OAO}
\subsubsection{Reliability Analysis}
Occlusion-Aware Offset (OAO) addresses the limitation of traditional IoU-based association in occluded scenarios, and this situation is analyzed in \cref{sec: 3}. Under occlusion, the spatial relationship between detections and tracks becomes unreliable due to:
\begin{itemize}
\item \textbf{Unreliable prediction of lost trajectory:} Multiple trajectories competing for the same detection region due to overlapping occlusions, as shown in \cref{fig:vis} (in \cref{sec: vis}).
\item \textbf{Positional ambiguity:} Occluded objects experience centroid shifts, bounding box distortions or strong competition, making IoU an inconsistent measure of true spatial proximity. This situation leads to cost confusion between detected bounding box and trajectories, causing ID switch as shown in \cref{fig:vis2} (in \cref{sec: vis}).
\end{itemize}

OAO mitigates these issues by incorporating occlusion awareness into the association cost matrix from the view of trajectory, effectively enhancing the discriminative power of the cost.

\subsubsection{Pseudocode}
We assume that the $N$ detections $\mathcal{D} \in \mathbb{R}^{N\times4}$ need to be associated with $M$ estimations $\mathbf{X}$. The pseudo-code for Occlusion-Aware Offset (OAO) is shown \cref{alg:oao}. The spatial consistency score about detections and estimations will be updated. Notably, the initial calculations of the spatial consistency score are different in different trackers. For example, Hybrid-SORT~\cite{yang2024hybrid} uses $\mathit{HMIoU}$; PD-SORT \cite{wang2025pd} uses DVIoU that is a 3D IoU. Herein, $C_{\mathit{IoU}}$ and $\mathit{IoU}$ are used uniformly.
\begin{algorithm}[t]
	\caption{Occlusion-Aware Offset \\ $\mathit{S} \leftarrow OAO(\mathbf{X}, \mathit{IoU}_{\mathcal{D}, \mathbf{X}})$)}
	\label{alg:oao}
	\textbf{Input}: estimations $\mathbf{X} \in \mathbb{R}^{M\times4}$ from tracked trajectories; $\mathit{IoU}_{\mathcal{D}, \mathbf{X}} \in \mathbb{R}^{N\times M}$ between $\mathcal{D}$ and $\mathbf{X}$\\
	\textbf{Parameter}: Coefficient $\tau$ that is used to balance IoU and $\hat{\mathit{Oc}}$ \\
	\textbf{Output}: Spatial consistency $\mathit{S}$
	\begin{algorithmic}[1]
		\STATE $\hat{\mathit{Oc}}^{\mathbf{X}} \in \mathbb{R}^{1\times M} \leftarrow OAM(\mathbf{X})$
		\STATE $\mathit{S} \in \mathbb{R}^{N \times M} \leftarrow \tau \cdot (1 - \hat{\mathit{Oc}}^{\mathbf{X}})[None, :] + (1 - \tau) \cdot \mathit{IoU}_{\mathcal{D}, \mathbf{X}}$ \COMMENT{This describes \cref{eq: COST_final}.}
		\STATE \textbf{return} $\mathit{S}$
	\end{algorithmic}
\end{algorithm}

\subsection{Bias-Aware Momentum}
\label{BAM}
\subsubsection{Reliability Analysis}
In theory, occlusion-aware tracking requires dynamic adjustment of both the process noise $\mathbf{Q}$ and measurement noise $\mathbf{R}$. However, real-time estimation of these parameters presents significant challenges:
\begin{itemize}
\item \textbf{Measurement noise estimation:} The error variance of the detection model under varying occlusion intensities is difficult to characterize. Occlusion-induced errors—including feature confusion and information loss—are varied and difficult to model accurately.
\item \textbf{Process noise estimation:} Motion model errors stem from natural scene dynamics and model residuals, which are inherently unobservable without ground truth. Furthermore, MOT in 2D images involves discrete observations of inherently nonlinear and irregular target motions. Any predictor necessarily exhibits lag during motion transitions, representing an unavoidable inherent error.
\end{itemize}

Herein, the additional experiments for BAM are conducted, as shown in \cref{tab: Refine-K}, for Kalman gain ($\mathbf{K}$) that can reflect and consider $\mathbf{Q}$ and $\mathbf{R}$.
\begin{table}[t]
\small
\centering
\begin{tabular}{cc|cccc}
\toprule
Refine $\mathbf{K}$     &   Dataset                   & MOTA          &   HOTA           &  AssA  & IDF1    \\
\midrule
                        & \multirow{2}{*}{DanceTrack} & 89.73         &   61.49          &  48.03 & 63.65   \\
\checkmark              &                             & 89.75         &   60.81          &  46.96 & 62.74   \\
\midrule
                        & \multirow{2}{*}{MOT17}      & 76.29         &   67.06          & 68.74  & 78.19    \\
\checkmark              &                             & 76.11         &   66.98          & 68.67  & 77.74    \\
\bottomrule
\end{tabular}
\caption{The influence of refining $\mathbf{K}$ through BAM (OA-SORT).}
\label{tab: Refine-K}
\end{table}

The result shows that BAM implements an instantaneous and implicit measurement reliability adjustment for low-confidence detections by directly modifying observation. This approach attempts to regulate the influence of new observations during the update step, eliminating the need for explicit error modeling or covariance recomputation—aligning well with both practical constraints and robustness objectives in real-world MOT. According to \cref{eq: bam} and \cref{eq: z_t}, the state update in \cref{eq: z_t} can be reformulated as:
\begin{equation}
	\mathbf{X}_{t|t} = \mathbf{X}_{t|t-1} + \mathrm{BAM} \cdot \mathbf{K}_t(\mathbf{Z}_t - \mathbf{H}_t \mathbf{X}_{t|t-1}).
\end{equation}

\subsubsection{Pseudocode}
We assume that a trajectory is association with a low-score detection, $d \in \mathbb{R}^{4}$; the trajectory's estimation $\mathbf{x}_{t|t-1} \in \mathbb{R}^{4}$. OAM is used on the process of updating combing with the occlusion coefficient of the latest observation $\hat{\mathit{Oc}}^{\mathbf{z}}$ of the trajectory. The pseudocode is shown in \cref{alg:bam}.
\begin{algorithm}[t]
	\caption{Bias-Aware Momentum ($\mathbf{x}_{t|t}=BAM(d, \mathbf{x}_{t|t-1}, \hat{\mathit{Oc}}^{\mathbf{z}}$)}
	\label{alg:bam}
	\textbf{Input}: $\mathit{det}$ that is detected bounding box;  $\mathbf{x}_{t|t-1}$; $\hat{\mathit{Oc}}^{\mathbf{z}}$\\
	\textbf{Output}: Optimized $\mathbf{x}_{t|t}$
	\begin{algorithmic}[1]
		\STATE $\mathit{IoU}_{\mathit{det}, \mathbf{x}_{t|t-1}} \in (0,1] \leftarrow C_{IoU}(\mathit{det}, \mathbf{x}_{t|t-1})$
		\STATE $BAM \leftarrow \mathit{IoU}_{\mathit{det}, \mathbf{x}_{t|t-1}} \cdot (1-\hat{\mathit{Oc}}^{\mathbf{z}})$ \COMMENT{This describes \cref{eq: BAM}.}
		\STATE $z' \leftarrow \mathit{det}$ \COMMENT{$z'$ denotes observations.}
		\STATE $\mathbf{x}_{t|t} \leftarrow \mathbf{x}_{t|t-1} + \mathit{BAM} \cdot \mathbf{K}_t(z' - \mathbf{H}_t \mathbf{x}_{t|t-1})$ \COMMENT{We omit unrelated Kalman filtering details for brevity. This describes \cref{eq: bam} and \cref{eq: z_t}.}
		\STATE \textbf{return} $\mathbf{x}_{t|t}$
	\end{algorithmic}
\end{algorithm}

\subsection{Association approach}
\label{ASS}
Given detections $\mathcal{D}$, trajectories $\mathcal{T}$, the Pseudo-codes for high-score detections $\mathcal{D}^{high}$, low-score detections $\mathcal{D}^{low}$ associations, and trajectory updating are shown as \cref{alg:high} and \cref{alg:low}, where $\mathcal{D}^{high} \cup \mathcal{D}^{low} = \mathcal{D}$ and $\mathcal{D}^{high} \cap \mathcal{D}^{low} = \emptyset$.

\begin{algorithm}[t]
	\caption{Association approach - high-score detections association\\ \{This step can be mixed with the low-score association, such as TrackTrack \cite{shim2025focusing}\}}
	\label{alg:high}
	\textbf{Input}: $\mathcal{D}^{high}, \mathcal{T}$
	\begin{algorithmic}[1]
		\STATE $\mathcal{E} \leftarrow$ [$t.estimation$ in $t$ for $\mathcal{T}$] \COMMENT{$\mathcal{E}$ indicate the estimation of trajectories.}
		\STATE $IoU_{\mathcal{D}^{high}, \mathcal{E}} \leftarrow C_{IoU}(\mathcal{D}^{high}, \mathcal{E})$
		\STATE $S \leftarrow OAO(\mathcal{E}, IoU_{\mathcal{D}^{high}, \mathcal{E}})$ \COMMENT{This describes \cref{sec: OAO}.}
		\STATE $S \leftarrow S + Other_{score}$ \COMMENT{Herein, other scores ($Other_{score}$) are used for allocation besides IoU. For example, Hybrid-SORT \cite{yang2024hybrid} and ByteTrack \cite{zhang2022bytetrack} use direction score and detection score. If using other scores, please calculate based on $S$. That's right here. Notably, this step can be mixed with the third line.}
		\STATE $high\_Match_{det}, high\_Match_{tra} \leftarrow$ the Hungarian Algorithm uses $S$ to allocation
	\end{algorithmic}
\end{algorithm}

\begin{algorithm}[t]
	\caption{Association approach - low-score detections association and trajectory updating}
	\label{alg:low}
	\textbf{Input}: $\mathcal{D}^{low}, \mathcal{T}^{un}\in \mathcal{T}$ 
	\begin{algorithmic}[1]
		\STATE $\mathcal{E}^{un} \leftarrow$ [$t.estimation$ in $t$ for $\mathcal{T}^{un}$] \COMMENT{$\mathcal{E}^{un}$ indicate the estimation of unmatched trajectories.}
		\STATE $IoU_{\mathcal{D}^{low}, \mathcal{E}^{un}} \leftarrow C_{IoU}(\mathcal{D}^{low}, \mathcal{E}^{un})$ 
		\STATE $S \leftarrow IoU_{\mathcal{D}^{low}, \mathcal{E}^{un}} + Other_{score}$ \COMMENT{Herein, other scores ($Other_{score}$) used for allocation besides IoU. For example, Hybrid-SORT \cite{yang2024hybrid} and ByteTrack \cite{zhang2022bytetrack} use direction score and detection score. If using other scores, please calculate based on $S$. That's right here.}
		\STATE $Low\_Match_{det}, Low\_Match_{tra} \leftarrow$ the Hungarian Algorithm uses $S$ to allocation \COMMENT{Store index in $\mathcal{D}$ and $\mathcal{T}$ of matched detections and trajectories.}
		\STATE $Match_{det} \leftarrow high\_Match_{det} + Low\_Match_{det}, Match_{tra} \leftarrow high\_Match_{tra} +Low\_Match_{tra}$ \COMMENT{If other matches exist, they are merged accordingly.}
		\\ \# -------------------\textbf{\textit{Trajectory updating}}------------------
		\FOR {each $d$, $t$ in $Match_{det}, Match_{tra}$}
		\IF {$\mathcal{D}[d]$ is in $\mathcal{D}^{low}$} 
		\STATE $\mathit{offset} \leftarrow BAM(\mathcal{D}[d], \mathcal{E}[t], \mathcal{T}[t].\hat{\mathit{Oc}})$  \COMMENT{$BAM$ is for $\mathcal{D}^{low}$. This describes \cref{eq: bam}.}
		\ELSE
		\STATE $\mathit{offset} \leftarrow 1$
		\ENDIF
		\STATE $\mathcal{T}[t] \leftarrow$ update $\mathcal{T}[t]$ using $\mathit{offset}$ and $\mathcal{D}[d]$
		\ENDFOR
		\STATE $\mathbf{Z} \leftarrow$ [$t.last\_observation$ for $t$ in $\mathcal{T}[Match_{tra}]$]
		\STATE $\hat{\mathit{Oc}}^{\mathbf{Z}} \leftarrow OAM(\mathbf{Z})$
		\STATE [$\mathcal{T}[t].\hat{\mathit{Oc}} \leftarrow \hat{\mathit{Oc}}[t]$ for $\hat{\mathit{Oc}}^{\mathbf{Z}}[t]$ in $\hat{\mathit{Oc}}^{\mathbf{Z}}$]
	\end{algorithmic}
\end{algorithm}

\section{Additional Experiments}
\subsection{$\sigma^x$ and $\sigma^y$ in GM}
\label{sec:gm}
For object $i$, $\sigma_i^x$ and $\sigma_i^y$ in GM can be indicated as $\frac{w_i}{k_x}$ and $\frac{h_i}{k_y}$. The performance results for different $k_x$ and $k_y$ values are presented as shown in \cref{tab: HOTAgm}, \cref{tab: AssAgm} and \cref{tab: IDF1gm}, where the rows represent the same $k_x$ value and the columns represent the same $k_y$ value. As a result, the $(k_x, k_y)=(5, 3)$ can achieve the best performance, \ie, $\sigma_i^x=\frac{w_i}{5}$ and $\sigma_i^y=\frac{h_i}{3}$.
\begin{table}[t]
	\small
	\centering
	\begin{tabular}{c|ccccc}
		\toprule
		&2&3&4&5&6 \\
		\midrule
		2	&60.357&60.477&60.366&60.49&60.836\\
		3	&60.543&	60.699&60.773&61.076&60.979\\
		4	&60.682&61.295&61.117&61.003&60.899\\
		5	&61.451&\textbf{61.667}&60.965&60.819&60.607\\
		6	&61.415&61.035&60.754&60.249&60.391\\
		\bottomrule
	\end{tabular}
	\caption{HOTA (\%) on the DanceTrack validation set under different $(k_x, k_y)$.}
	\label{tab: HOTAgm}
\end{table}
\begin{table}[t]
	\small
	\centering
	\begin{tabular}{c|ccccc}
		\toprule
		&2&3&4&5&6 \\
		\midrule
		2&46.171&46.326&46.268&46.351&47.023\\
		3&46.494&46.794&46.902&47.374&47.363\\
		4&46.732&47.731&47.418&47.318&47.167\\
		5&47.932&\textbf{48.257}&47.211&47.074&46.719\\
		6&47.926&47.379&47.022&46.207&46.472\\
		\bottomrule
	\end{tabular}
	\caption{AssA (\%) in DanceTrack under different $k_x, k_y$.}
	\label{tab: AssAgm}
\end{table}
\begin{table}[t]
	\small
	\centering
	\begin{tabular}{c|ccccc}
		\toprule
		&2&3&4&5&6 \\
		\midrule
		2&61.712&61.848&62.176&62.249&62.924 \\
		3&62.031&62.312&62.83&63.348&63.037\\
		4&62.198&63.373&63.182&63.069&62.818\\
		5&63.444&\textbf{63.789}&63.074&62.778&62.507\\
		6&63.38&62.943&62.682&61.938&62.292\\
		\bottomrule
	\end{tabular}
	\caption{IDF1 (\%) in DanceTrack under different $k_x, k_y$.}
	\label{tab: IDF1gm}
\end{table}

\begin{figure}[t]
	\centering
	\includegraphics[width=\linewidth]{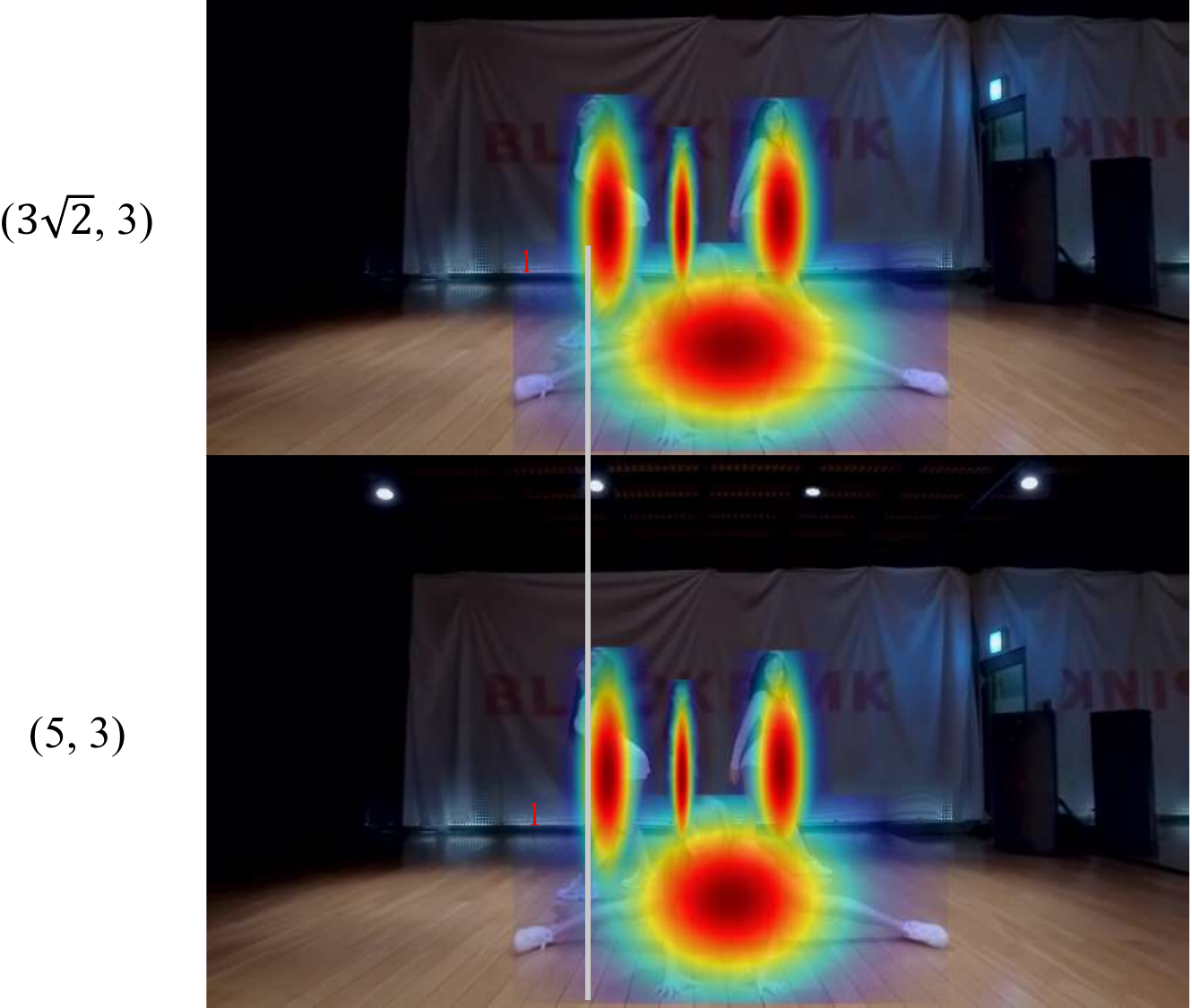}
	\caption{The example about different $(k_x, k_y)$.}
	\label{fig: GM}
\end{figure}

The result primarily shows the approximate global optimum on the DanceTrack validation set. However, when $k_x$ or $k_y$ is larger, the Gaussian distribution might be prone to disorder as shown in \cref{fig: GM}. To avoid particularity of $k_x$ and $k_y$, we attempted to reduce $k_x:k_y=3*\sqrt{2}: 3 = 1.414:1$, because the quadratic term $(\sigma^x_n)^2$ in \cref{eq: GM} as well as $k_x \in (4, 5)$. The results in \cref{tab: different} agree with us that the performance is better in the DanceTrack test dataset when $k_x = (3*\sqrt{2}, 3)$. Finally, $\sigma^x_n$ and $\sigma^y_n$ are set to $\frac{w_n}{3*\sqrt{2}}$ and $\frac{y_n}{3}$, respectively. 

\begin{table}[t]
\small
\centering
\begin{tabular}{cc|ccc}
\toprule
Dataset                             &  $(k_x, k_y)$           &  HOTA           &  AssA             & IDF1      \\
\midrule
\multirow{3}{*}{test}               &  -                      & 62.20           & 47.40             & 63.00    \\
                                    & $(3*\sqrt{2}, 3)$       & \textbf{63.13}  &  \textbf{48.53}   & 64.17    \\
                                    &     (5, 3)              & 62.93           & 48.23             & \textbf{64.23}    \\
\midrule
\multirow{3}{*}{val}                &  -                      & 59.39           &   44.90           &   60.67  \\
                                    & $(3*\sqrt{2}, 3)$       & 61.49           &  48.03            & 63.65    \\
                                    &     (5, 3)              & \textbf{61.67}  &  \textbf{48.26}   & \textbf{63.79}    \\
\midrule
\multirow{3}{*}{Average}            &    -                    & 60.80           &   46.15           &   61.84   \\
                                    & $(3*\sqrt{2}, 3)$       & \textbf{62.31}  &  \textbf{48.28}   & 63.91     \\
                                    &     (5, 3)              & 62.30           &  48.25            &\textbf{ 64.01}     \\
\bottomrule
\end{tabular}
\caption{The performance of OA-SORT under different $k_x$ in DanceTrack. '-' indicates baseline Hybrid-SORT.}
\label{tab: different}
\end{table}

\subsection{Ablation Study for MOT17}
\begin{figure}
\centering
\begin{subfigure}{\linewidth}
\centering
\includegraphics[width=0.48\linewidth]{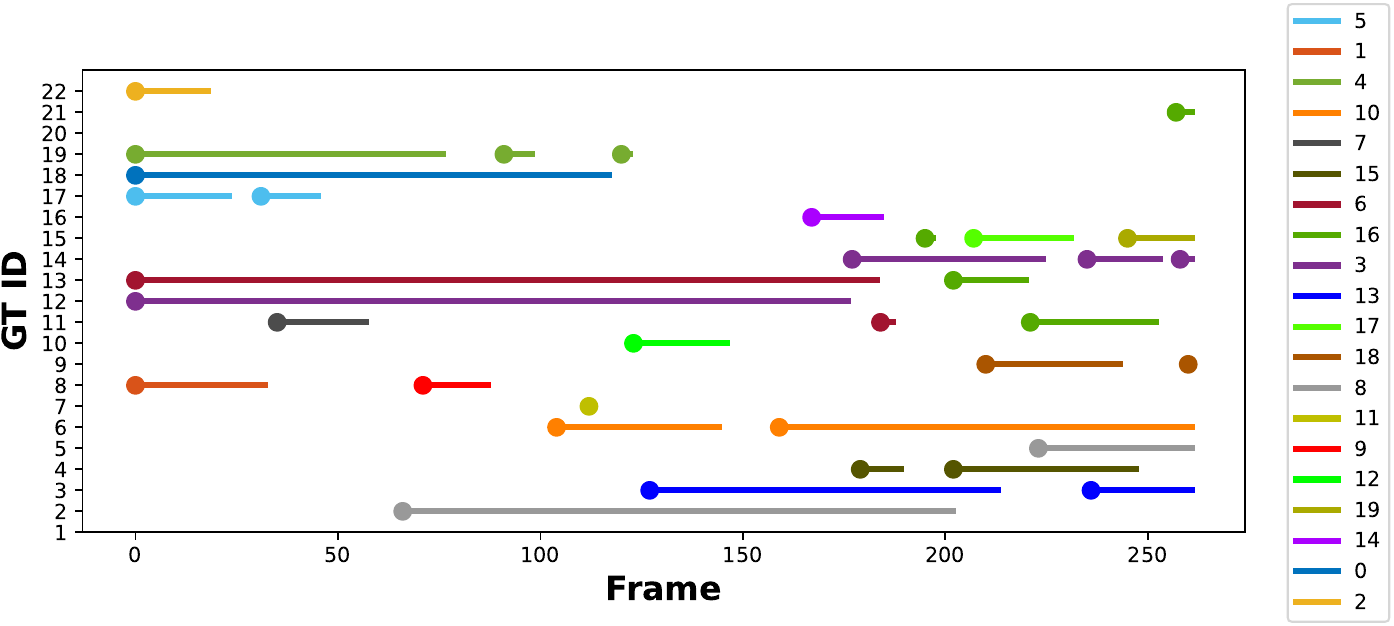} 
\includegraphics[width=0.48\linewidth]{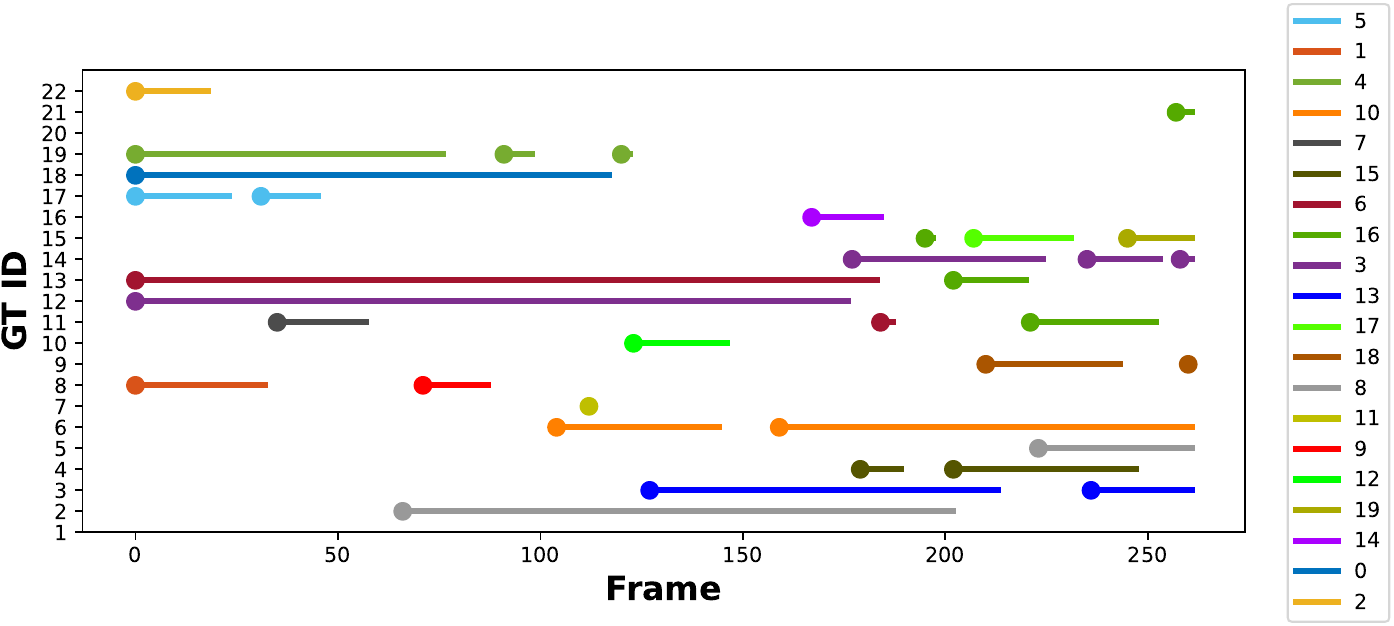}
\caption{The ID situation under detection provided by Host. The left is Hybrid-SORT's ID result, and the right is OA-SORT's result.}
\label{fig:MOT17Ablation-a}
\end{subfigure}
\hfill
\begin{subfigure}{\linewidth}
\centering
\includegraphics[width=0.48\linewidth]{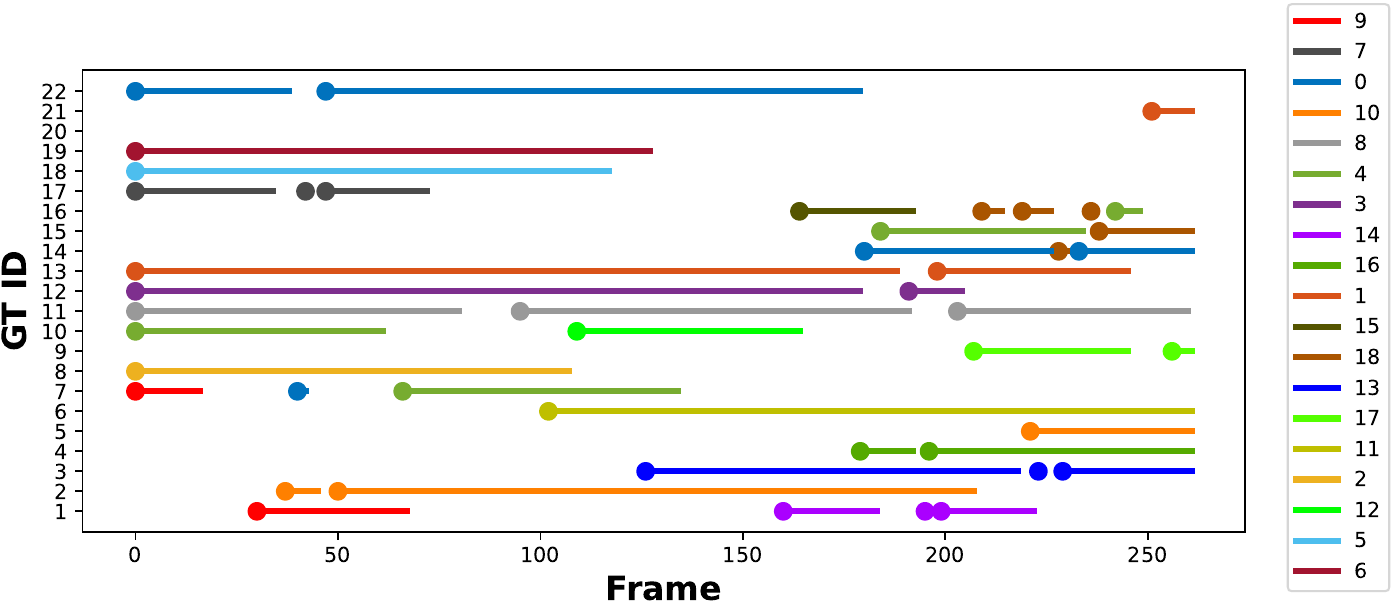}
\includegraphics[width=0.48\linewidth]{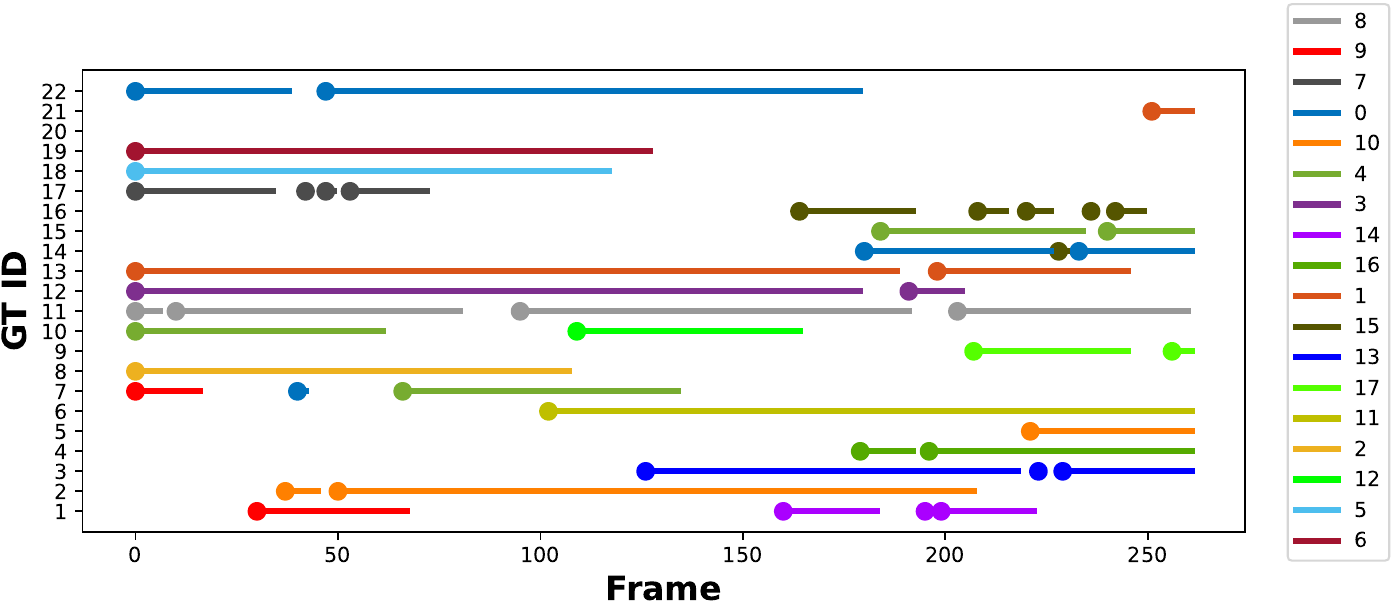}
\caption{The ID situation under detection provided by Hybrid-SORT. The left is Hybrid-SORT's ID result, and the right is OA-SORT's result.}
\label{fig:MOT17Ablation-b}
\end{subfigure}
\caption{The results on MOT17-09. The different color indicates different trajectory number and the dots on the line represent the starting point of trajectory interruption or identity change. GT ID represents the actual ID. The number in the legend represents the trajectory number rather than GT ID. The legend also reflects the total number of trajectories in the sequence.}
\label{fig:MOT17Ablation}
\end{figure}

The additional ablations for MOT17 are executed as shown in \cref{tab: componentMOT17}. The results exhibit the performance improvement from different detection, including YOLOX and Public detection (FRCNN) provided by Hybrid-SORT and host, respectively. Overall, in Public detection with lower detection quality, the occlusion-aware framework can still ensure the performance. However, the benefits it brings are not as good as those from high-quality detection. This is due to missed detections and false positives as shown in \cref{fig:MOT17Ablation}, which are not the focus of the occlusion-aware framework.
\begin{table}[t]
\small
\setlength{\tabcolsep}{1.2mm}
\centering
\begin{tabular}{cc|cccc}
\toprule
Detection                &   Method        & MOTA        &   HOTA         &  AssA & IDF1   \\
\midrule
\multirow{2}{*}{YOLOX}   & Hybrid-SORT     & 75.65       &   66.75       & 68.36  & 77.64   \\
                         & OA-SORT         & 76.29       &   67.06       & 68.74  & 78.19   \\
\midrule
\multirow{2}{*}{Public}  & Hybrid-SORT     &  46.55      &   50.16     & 59.49  & 56.89   \\
                         & OA-SORT         &  46.64      &   50.22     & 59.54  & 57.04   \\
\bottomrule 
\end{tabular}
\caption{Ablation under different detection on MOT17-val.}
\label{tab: componentMOT17}
\end{table}

Finally, the camera-motion compensation (CMC \cite{maggiolino2023deep}) is integrated into OA-SORT under the detection provided by Hybrid-SORT. The results are reported in \cref{tab: cmc}. Comparing enhancement brought by occlusion-awareness with or without CMC, we find that CMC brings benefits for occlusion-aware framework in HOTA. When CMC is not used, the HOTA and MOTA performance is improved by 0.31\% and 0.64\% from Hybrid-SORT to OA-SORT. In contrast, HOTA and MOTA enhancement with CMC is 0.39\% and 0.54\%.
\begin{table}[t]
\scriptsize
\setlength{\tabcolsep}{1mm}
\centering
\begin{tabular}{cc|ccccl}
\toprule
CMC                &   Method        & MOTA          &   HOTA           &  AssA  & IDF1    &  IDS \\
\midrule
                   & Hybrid-SORT     & 75.65         &   66.75          & 68.36  & 77.64   & 250  \\
\checkmark         & Hybrid-SORT     & 76.21         &   67.83          & 70.15  & 80.25   & 204 \\
\midrule
                   & OA-SORT         & 76.29 (+0.64) &   67.06  (+0.31) & 68.74  & 78.19   & 239 \\
\checkmark         & OA-SORT         & 76.75 (+0.54) &   68.22  (+0.39) & 70.73  & 80.93   & 170 \\
\bottomrule
\end{tabular}
\caption{The influence of CMC on MOT17-val.}
\label{tab: cmc}
\end{table}
\subsection{Additional data visualization under Occlusion}
\begin{figure}[t]
	\centering
	\begin{subfigure}{\linewidth}
		\includegraphics[width=\linewidth]{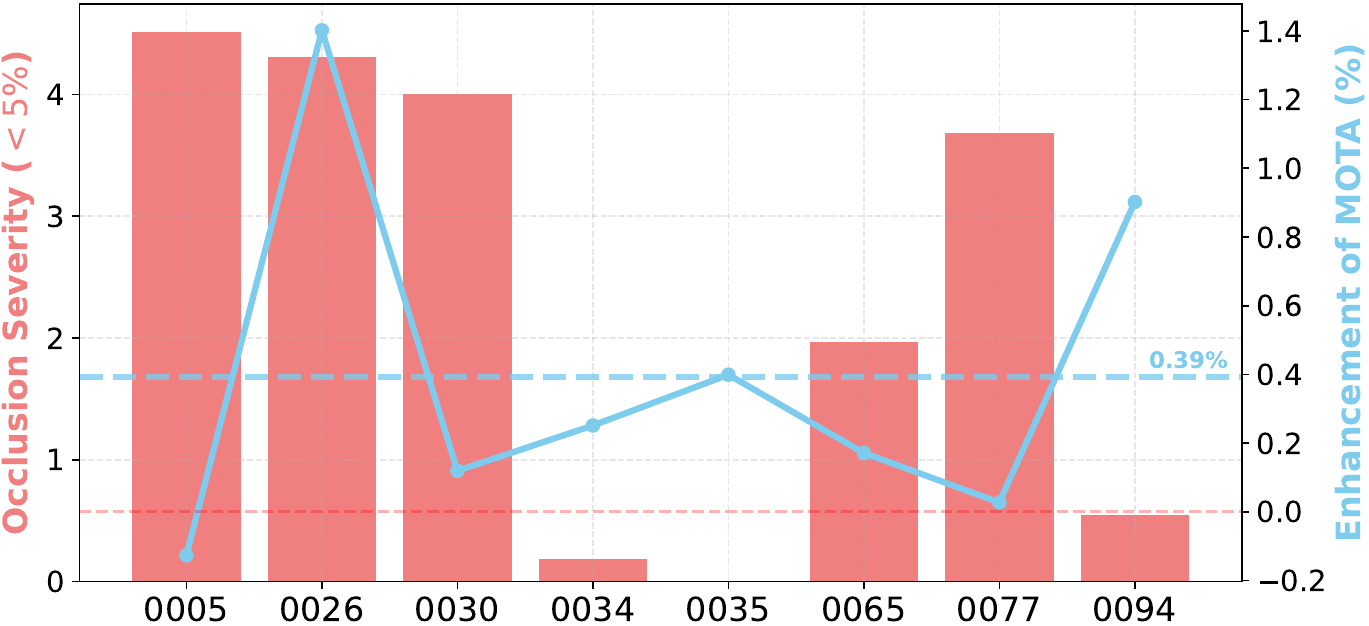} \\
		\includegraphics[width=\linewidth]{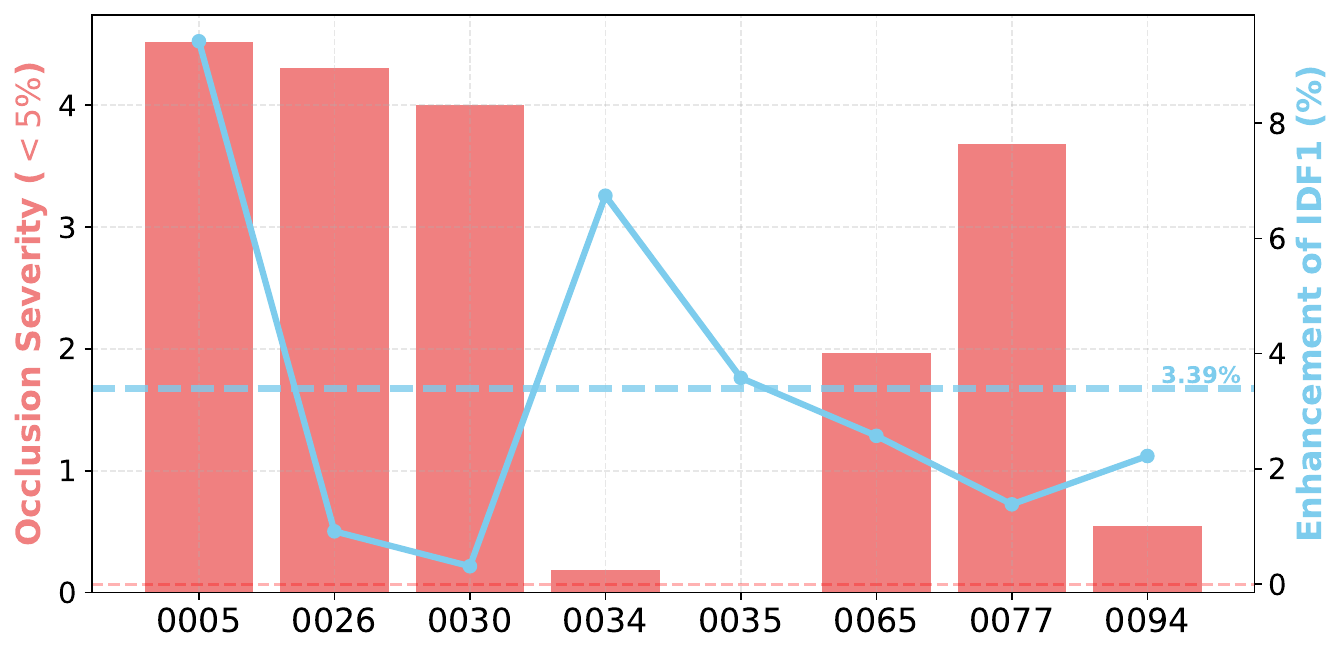}
		\label{fig:occlusion_add_<5}
	\end{subfigure}
	\hfill
	\begin{subfigure}{\linewidth}
		\includegraphics[width=\linewidth]{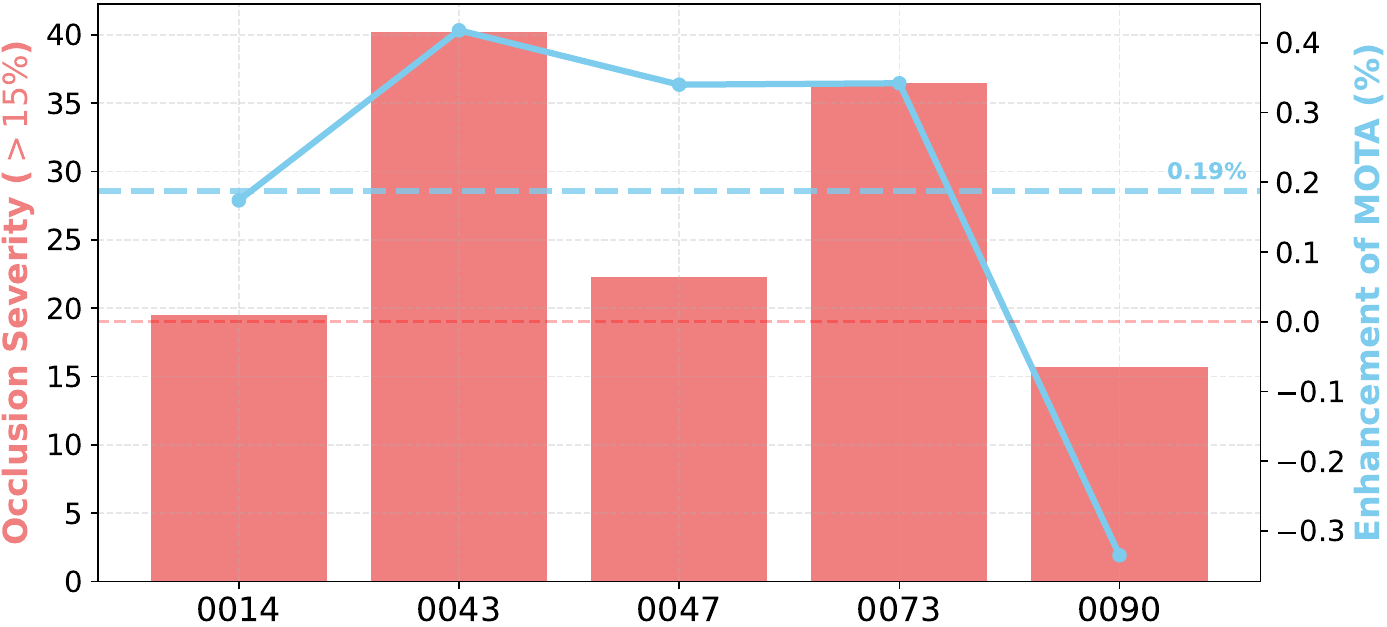} \\
		\includegraphics[width=\linewidth]{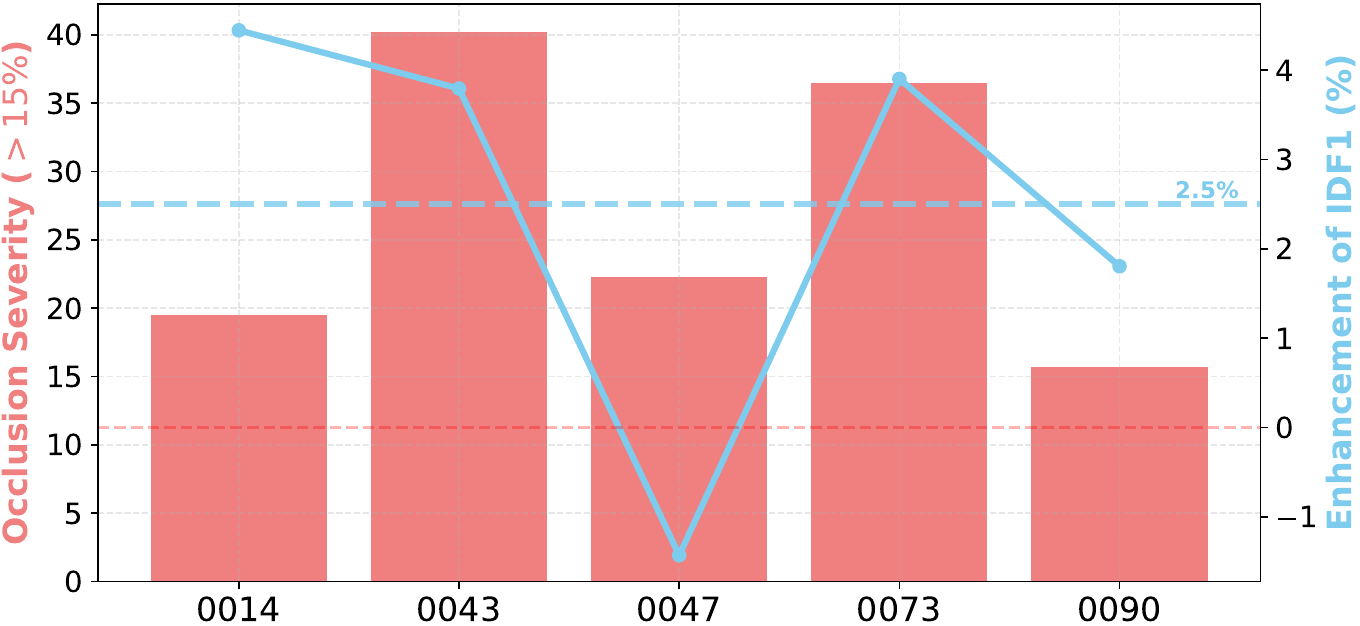}
		\label{fig:occlusion_add_>15}
	\end{subfigure}
	\caption{Performance improvement in MOTA and IDF1 over Baseline on the DanceTrack validation set under high and low occlusion severity.}
	\label{fig:occlusion_add}
\end{figure}

According to \cref{fig:occlusion}, the additional data, IDF1 and MOTA, are provided as shown in \cref{fig:occlusion_add}. Although the HOTA in the \#0026 sequence has decreased, both IDF1 and MOTA have improved, especially in terms of MOTA, which has increased by approximately 1.4\%. Combing \cref{fig:occlusion}, the results show that the main reason for this situation may be that the proposed occlusion-aware framework focuses on instantaneous states rather than long-term, resulting in unstable tracking.

\subsection{MOT20 test set}
\label{MOT20}
In \cref{tab: mot20}, the results under MOT20~\cite{dendorfer2020mot20} are presented. OA-SORT maintains good performance. Compared with Hybrid-SORT, OA-SORT improves with 0.4 IDF1 and 0.3 AssR. The results indicate that OAO and BAM can still maintain performance in dense-occlusion scenarios.
\begin{table*}[t]
	\centering
	\begin{tabular}{@{}lcccccccc}
		\toprule
		Method                   & HOTA (\%)$\uparrow$  & MOTA (\%)$\uparrow$ & IDF1 (\%)$\uparrow$ & FP ($10^4$)$\downarrow$ & FN ($10^4$)$\downarrow$ & IDs$\downarrow$ & AssA (\%)$\uparrow$ & AssR (\%)$\uparrow$\\
		\midrule
		FairMOT \cite{fairmot} & 54.6  & 61.8 & 67.3 &10.30&8.89&5,243&  54.7 & 60.7 \\
		CSTrack \cite{liang2022rethinking} & 54.0  & 66.6 & 68.6 &2.54&14.4&3,196&  54.0&57.6\\
		TransMOT \cite{chu2023transmot} & 61.9  & 77.5 & 75.2 &3.42&8.08&1,615&  60.1 & 66.3\\
		UTM \cite{you2023utm} & 62.5  & 74.3 & 79.8 & 3.00&8.15&1,228& - & -\\
		\rowcolor{blue2}
		GHOST \cite{seidenschwarz2023simple}                    & 61.2  & 73.7 & 75.2&-&-& 1,264 & -& -\\
		\rowcolor{blue2}
		ByteTrack \cite{zhang2022bytetrack}                 & 61.3  & \textbf{77.8} & 75.2&2.62&8.76& 1,223 & 59.6& 66.2 \\
		\rowcolor{blue2}
		OC-SORT \cite{cao2023observation}               & 62.1  & 75.5 & 75.9&\textbf{1.80}&10.8&\textbf{913} & 62.0& 67.5\\
		\rowcolor{blue2}
		StrongSORT \cite{du2023strongsort}            & 61.5  & 72.2 & 75.9 &-&-&1,066&  63.2 &- \\
		\rowcolor{blue2}
		AIPT \cite{zhang2024aipt}          & 62.1  & 76.9 & 75.4 & 2.10 &9.80&1,134&  61.1 &  -\\  
		\rowcolor{blue2}
		Hybrid-SORT \cite{yang2024hybrid}          & 62.5  & 76.4& 76.2 & 3.59 &8.50&1,300&  62.0 & 68.4\\  
		\rowcolor{blue2}
		OA-SORT & \textbf{62.6}  &  76.5 &  \textbf{76.6} &3.59&\textbf{8.46}&1,274&   \textbf{62.1}& \textbf{68.7} \\
		\bottomrule
	\end{tabular}
	\caption{Results on MOT20-test with the private detections.}
	\label{tab: mot20}
\end{table*}

\subsection{Visualization}
\label{sec: vis}
\begin{figure*}[t]
	\centering
	\begin{subfigure}{0.48\linewidth}
		\centering
		\includegraphics[width=\linewidth]{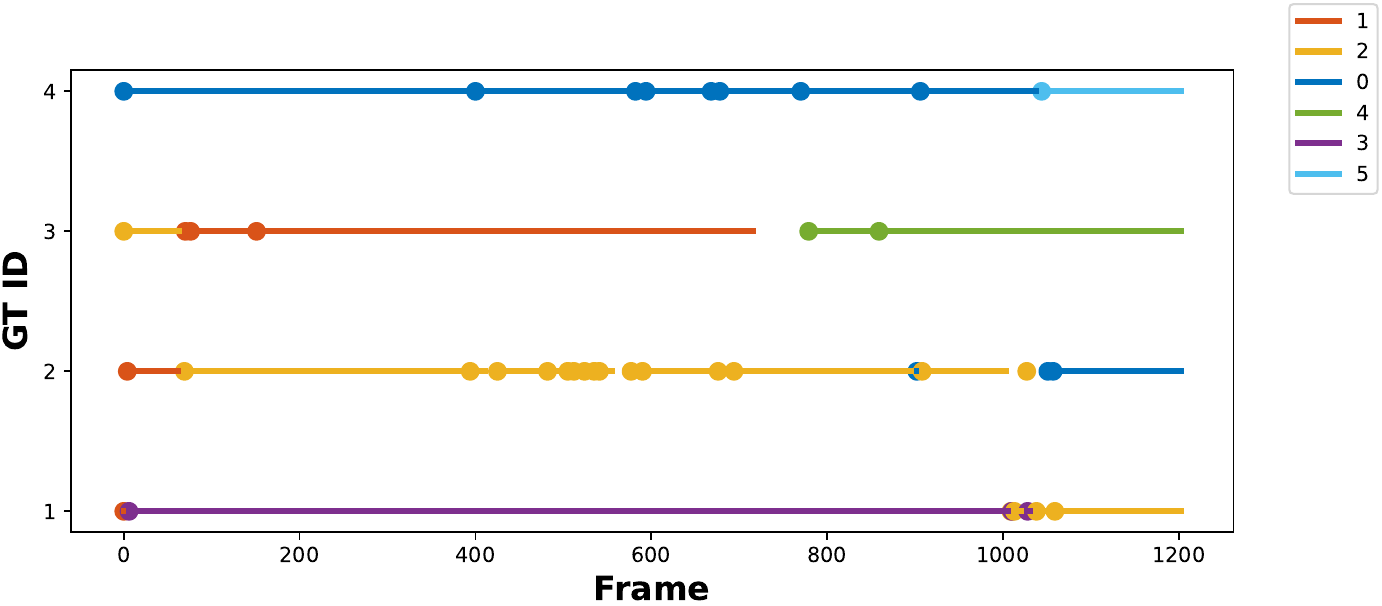} \\
		\includegraphics[width=\linewidth]{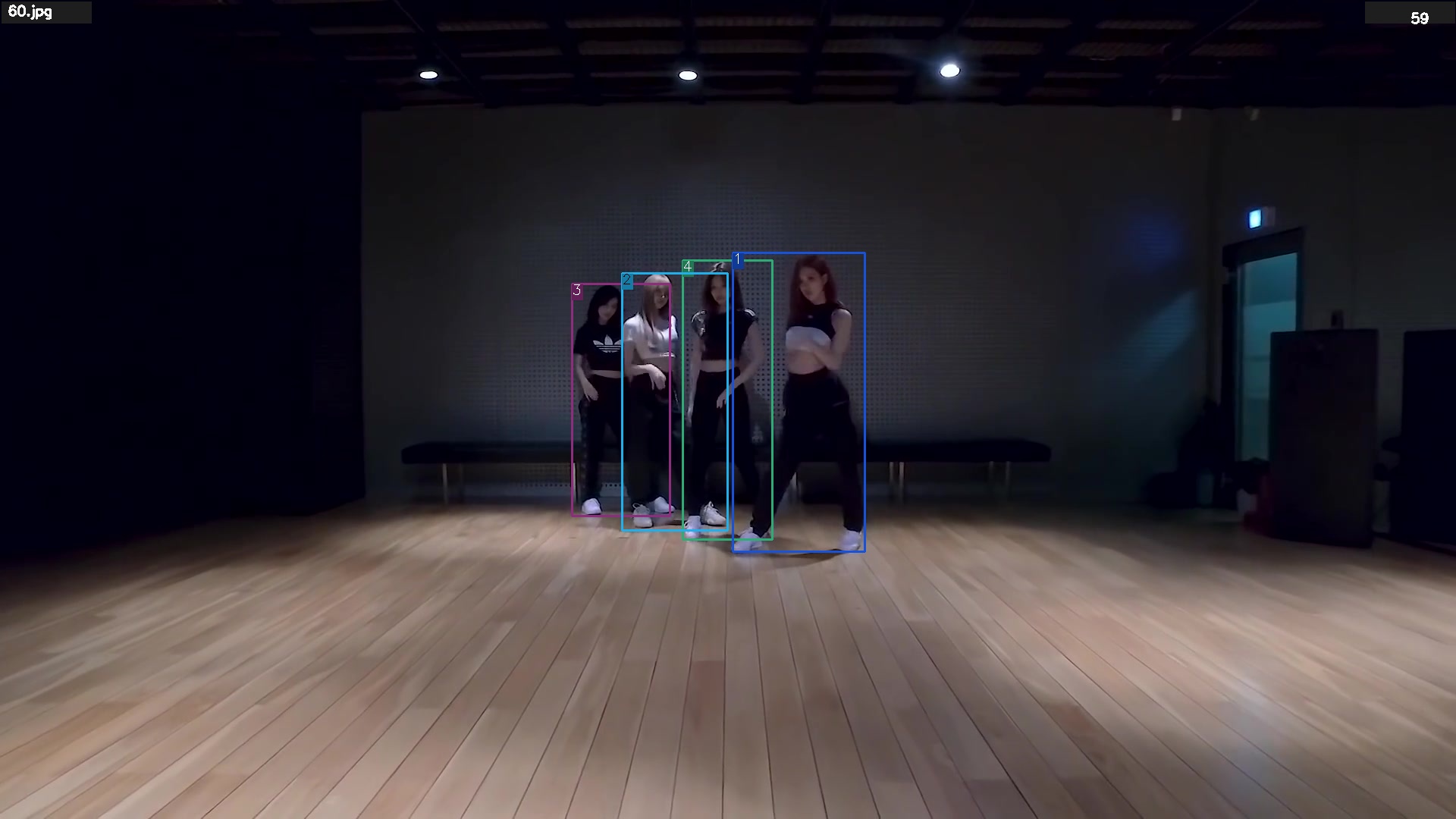} 
		\includegraphics[width=\linewidth]{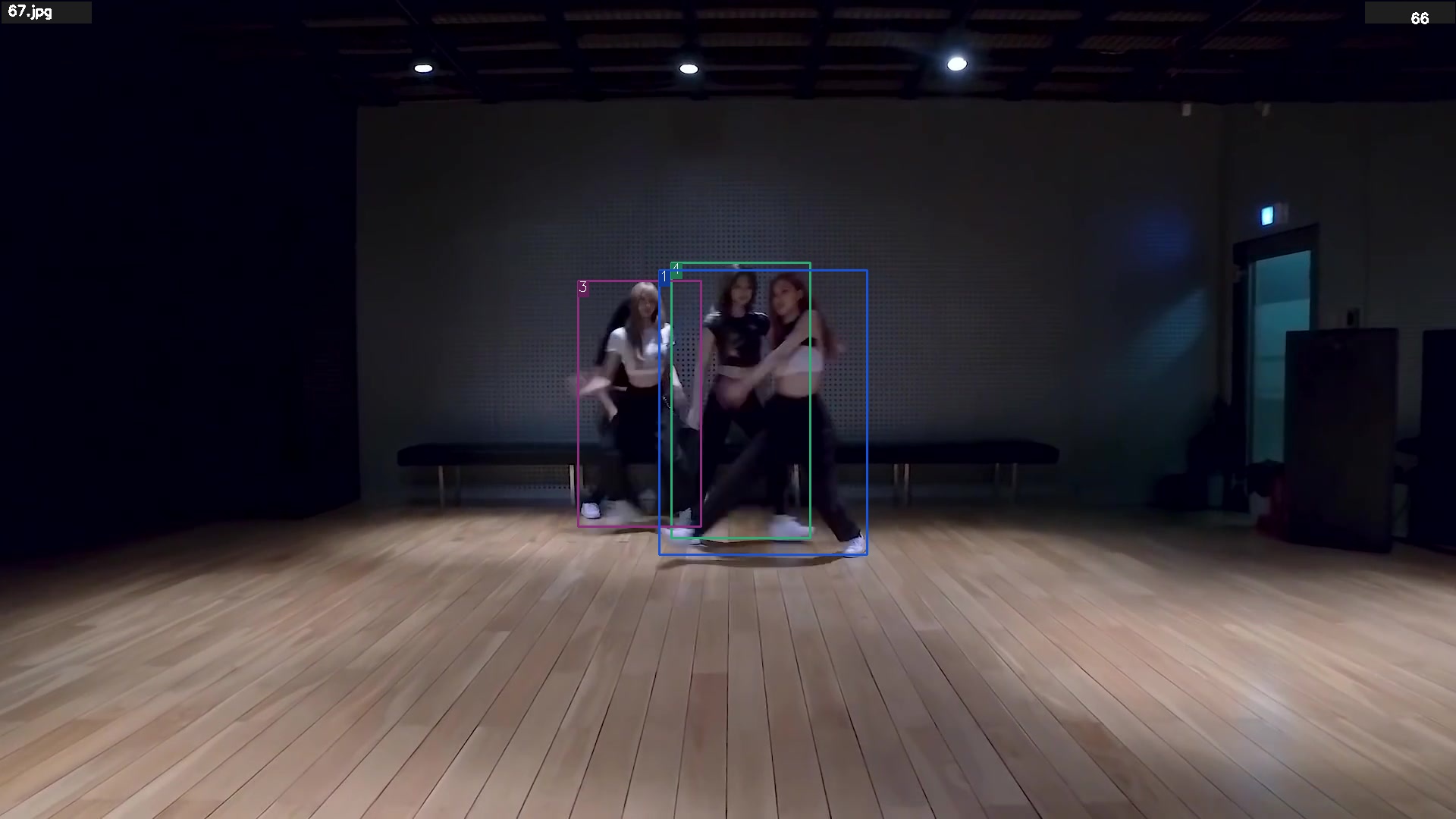} 
		\includegraphics[width=\linewidth]{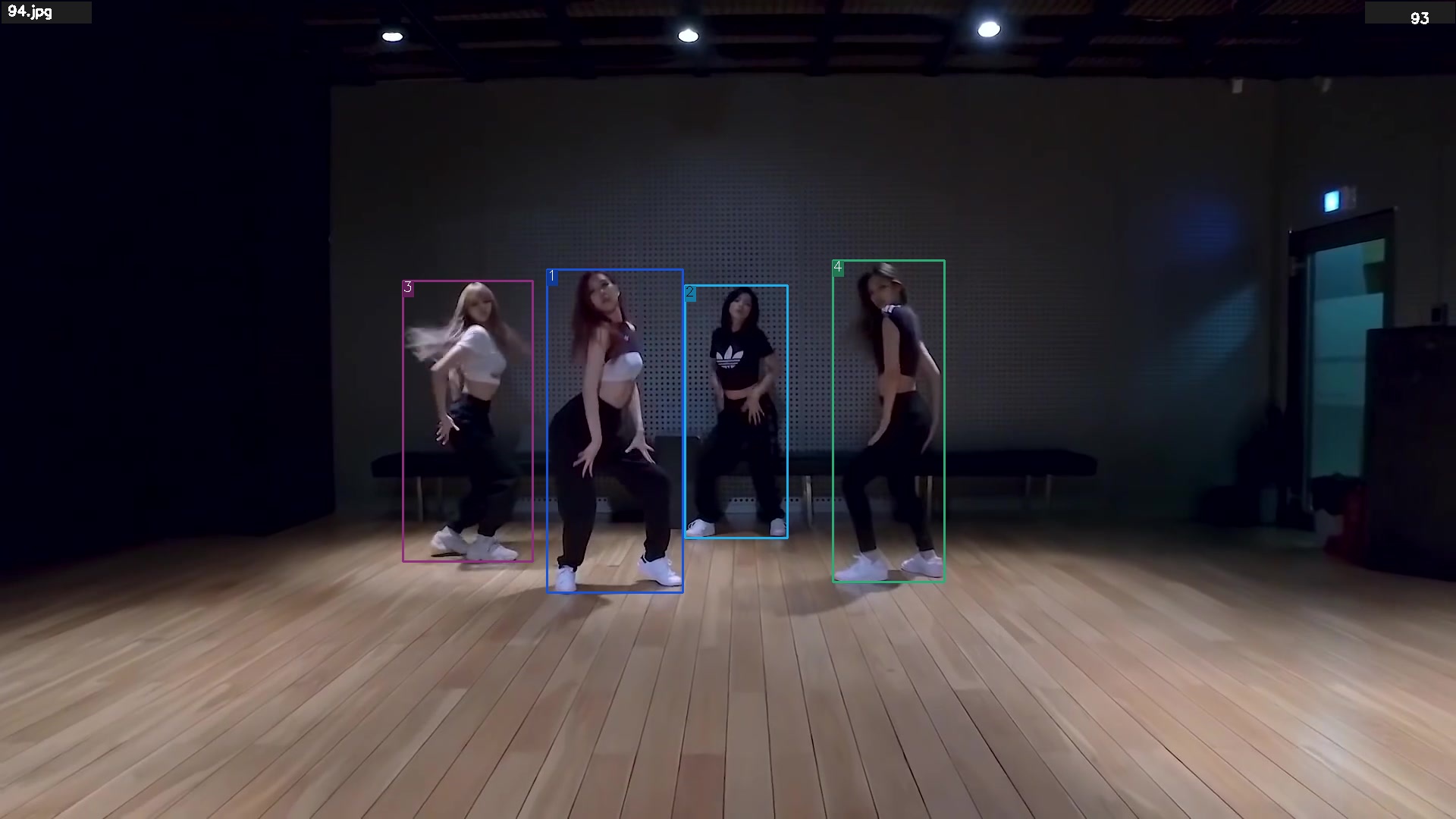} 
		\caption{The results through Hybrid-SORT.}
		\label{fig:vis-a}
	\end{subfigure}
	\hfill
	\begin{subfigure}{0.48\linewidth}
		\centering
		\includegraphics[width=\linewidth]{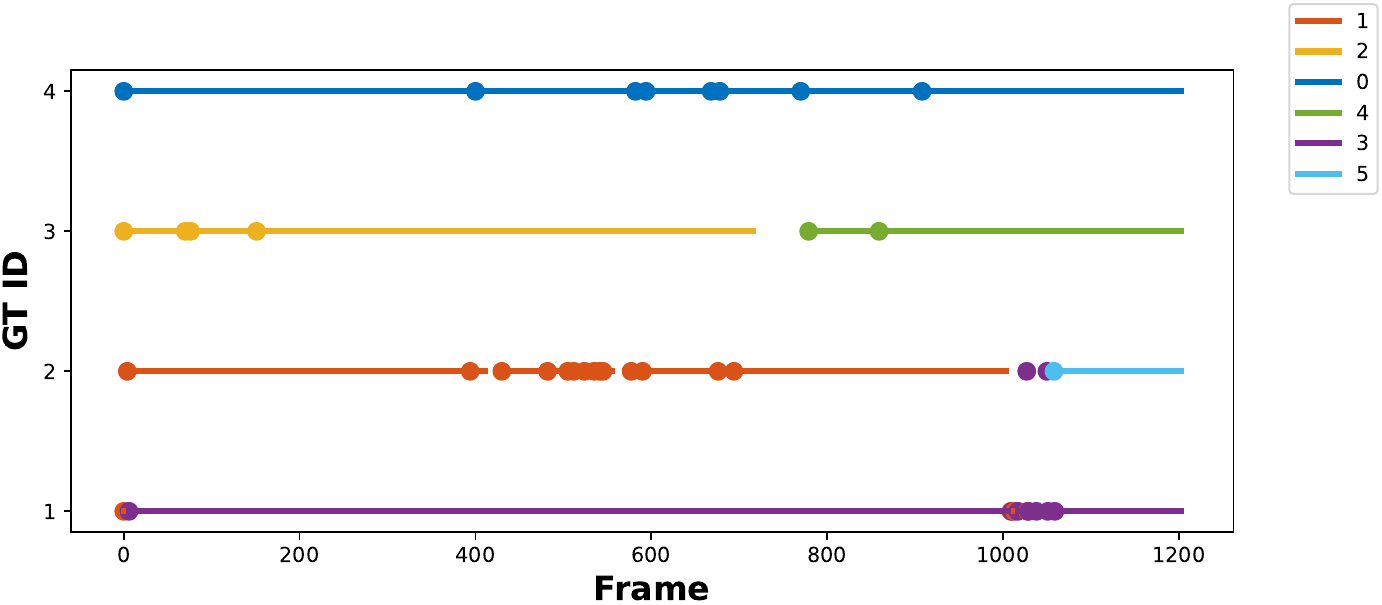}\\
		\includegraphics[width=\linewidth]{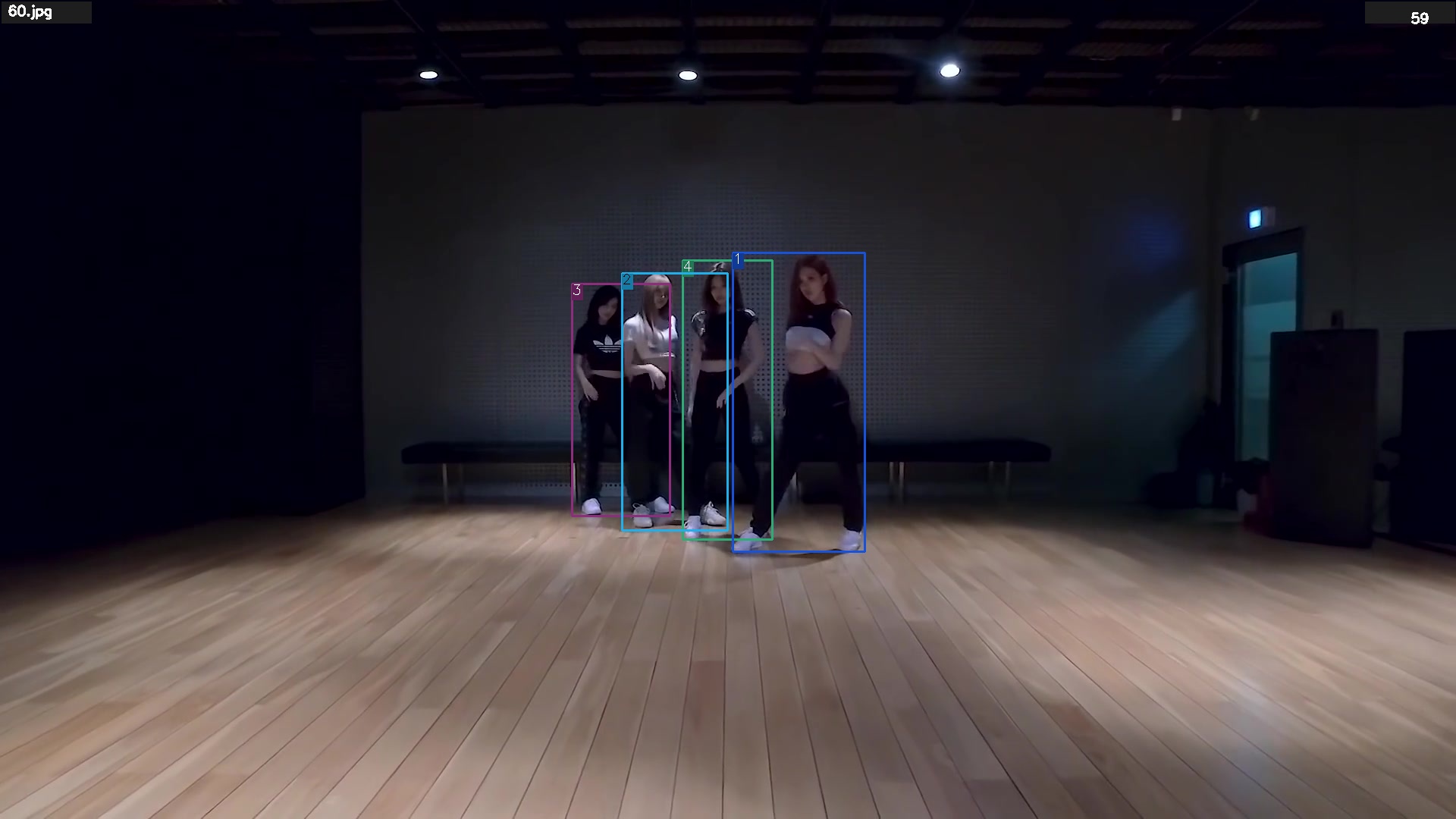} 
		\includegraphics[width=\linewidth]{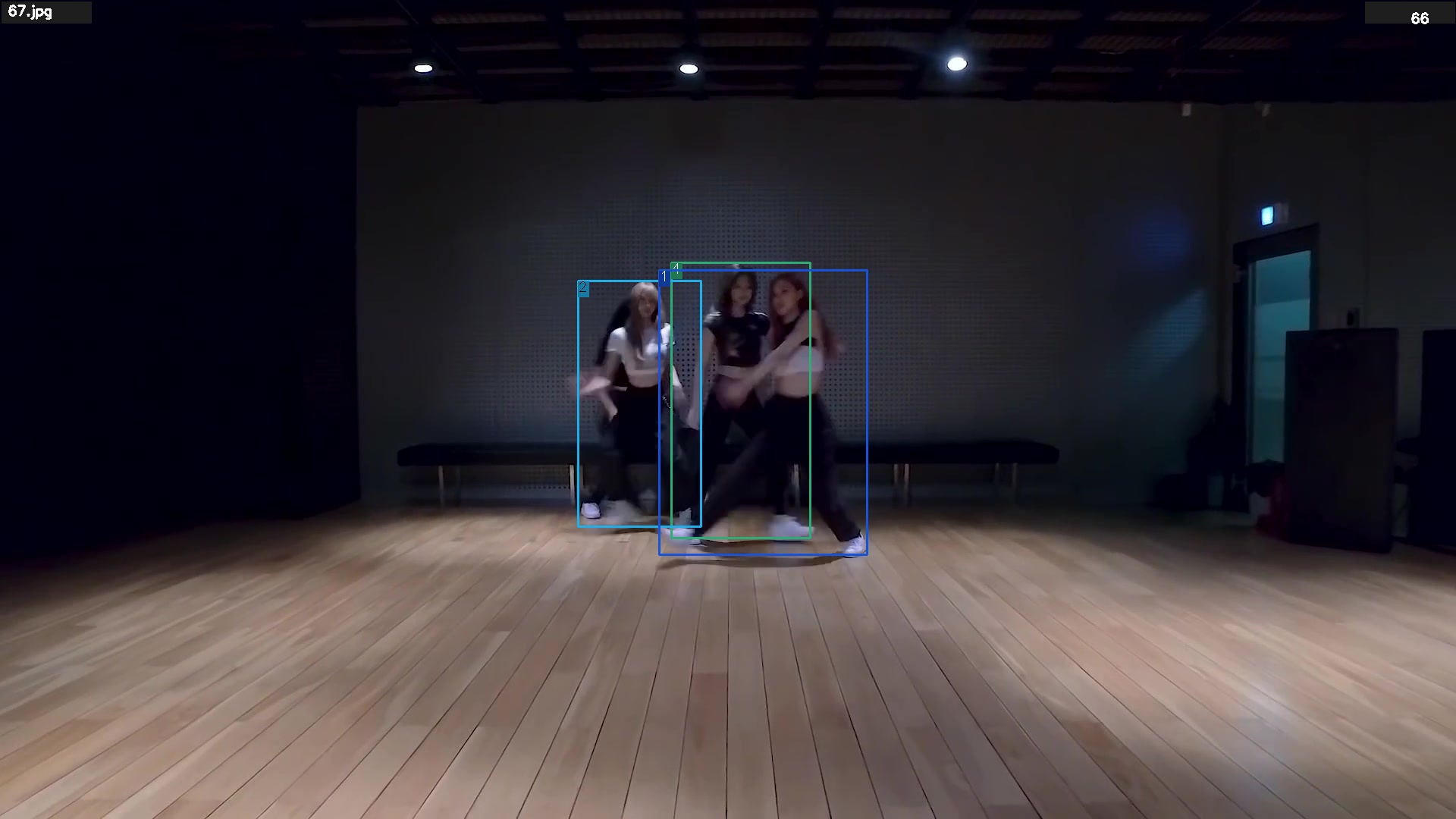} 
		\includegraphics[width=\linewidth]{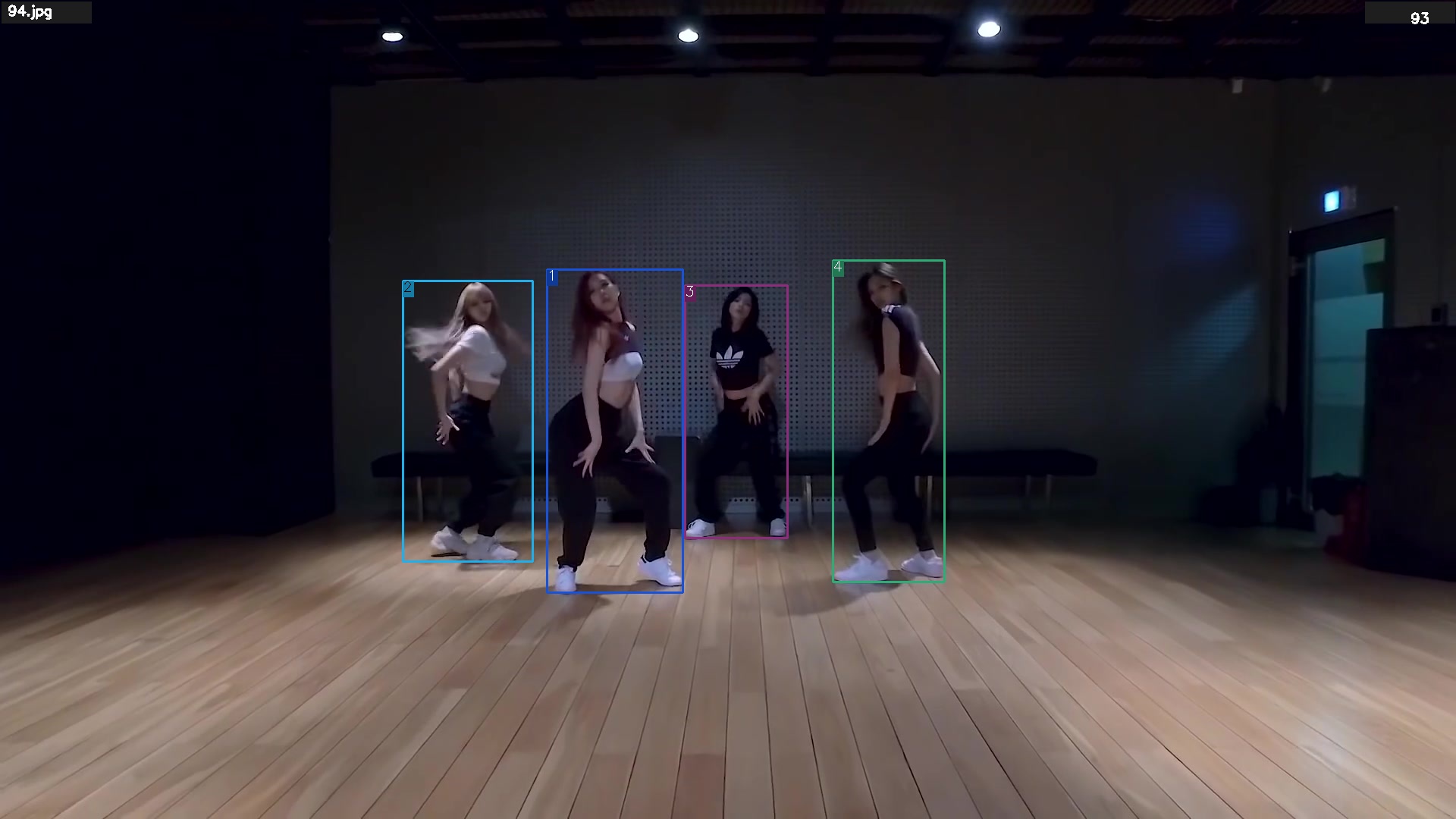} 
		\caption{The results through OA-SORT.}
		\label{fig:vis-b}
	\end{subfigure}
	\caption{The results on DanceTrack0058. The different color indicates different trajectory number and the dots on the line represent the starting point of trajectory interruption or identity change. GT ID represents the actual ID. The number in the legend represents the trajectory number rather than GT ID. The legend also reflects the total number of trajectories in the sequence.}
	\label{fig:vis}
\end{figure*}

\begin{figure*}[t]
	\centering
	\begin{subfigure}{0.48\linewidth}
		\includegraphics[width=\linewidth]{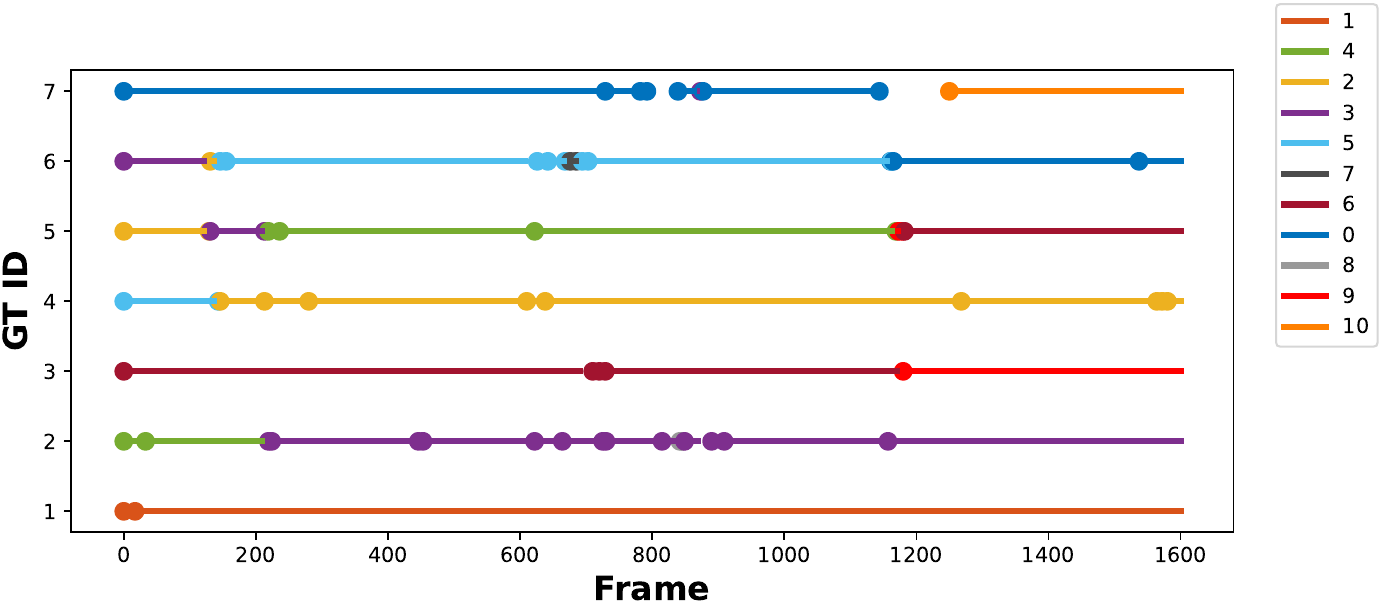} \\
		\includegraphics[width=\linewidth]{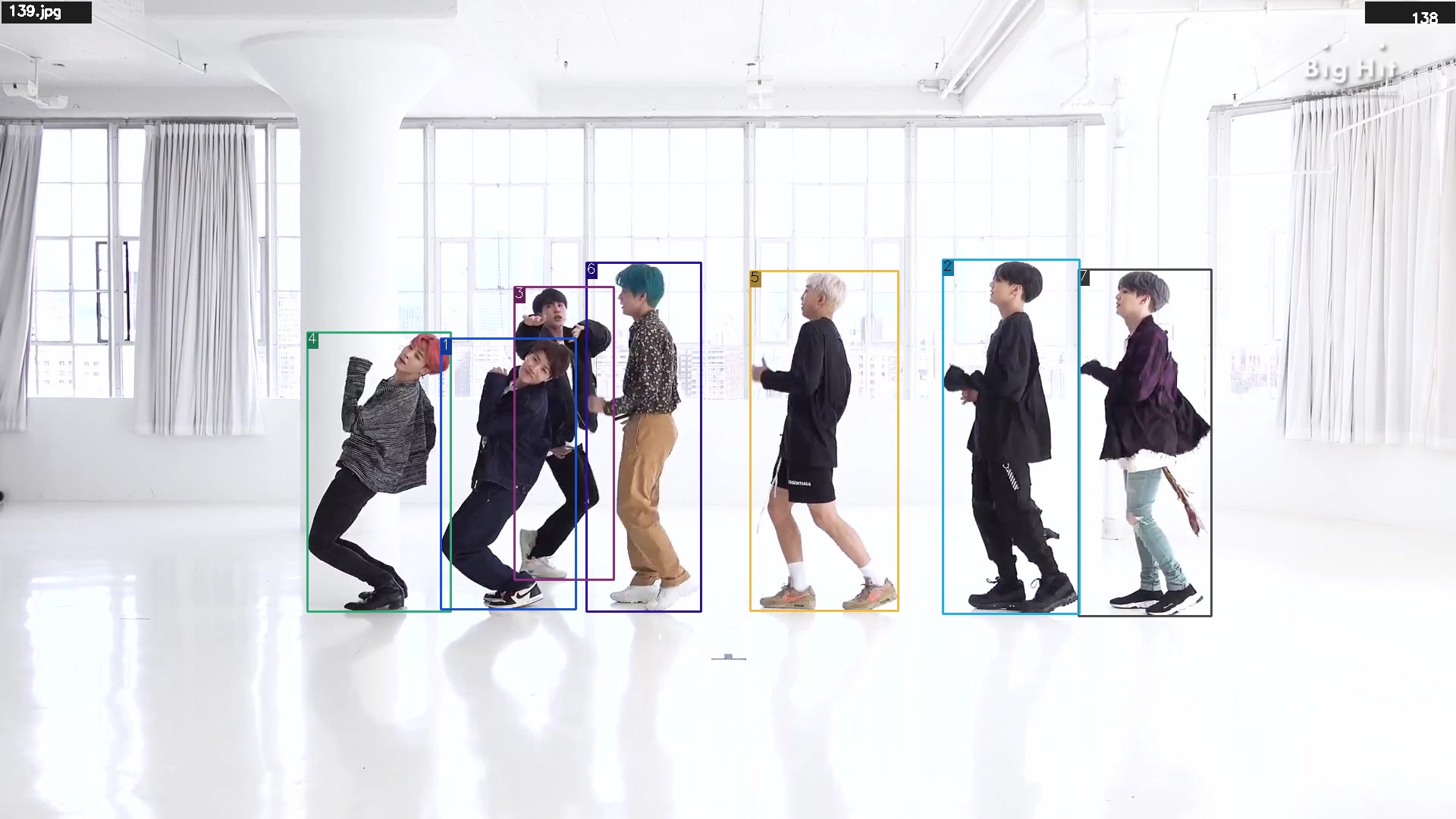} \\
		\includegraphics[width=\linewidth]{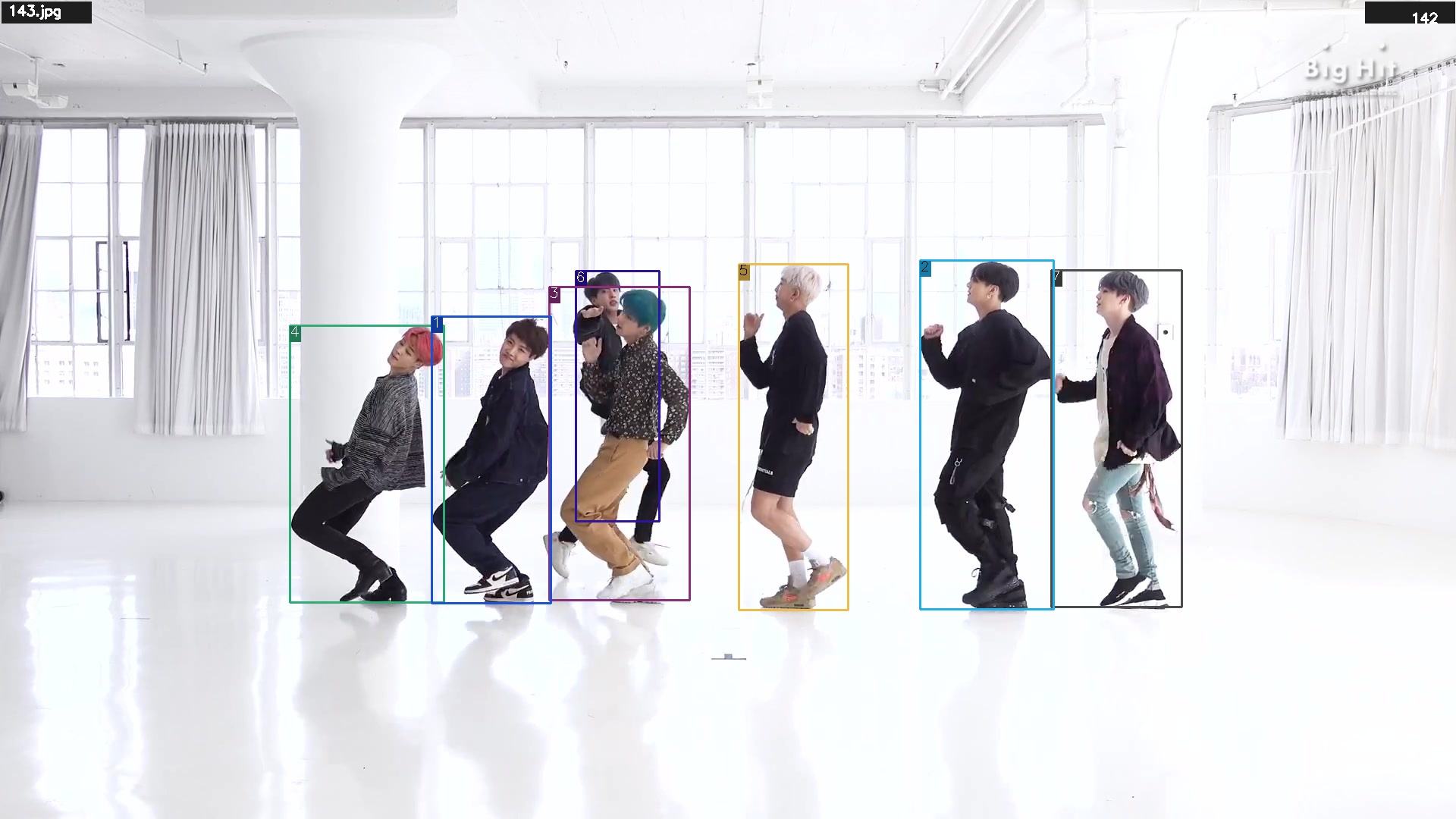} \\
		\includegraphics[width=\linewidth]{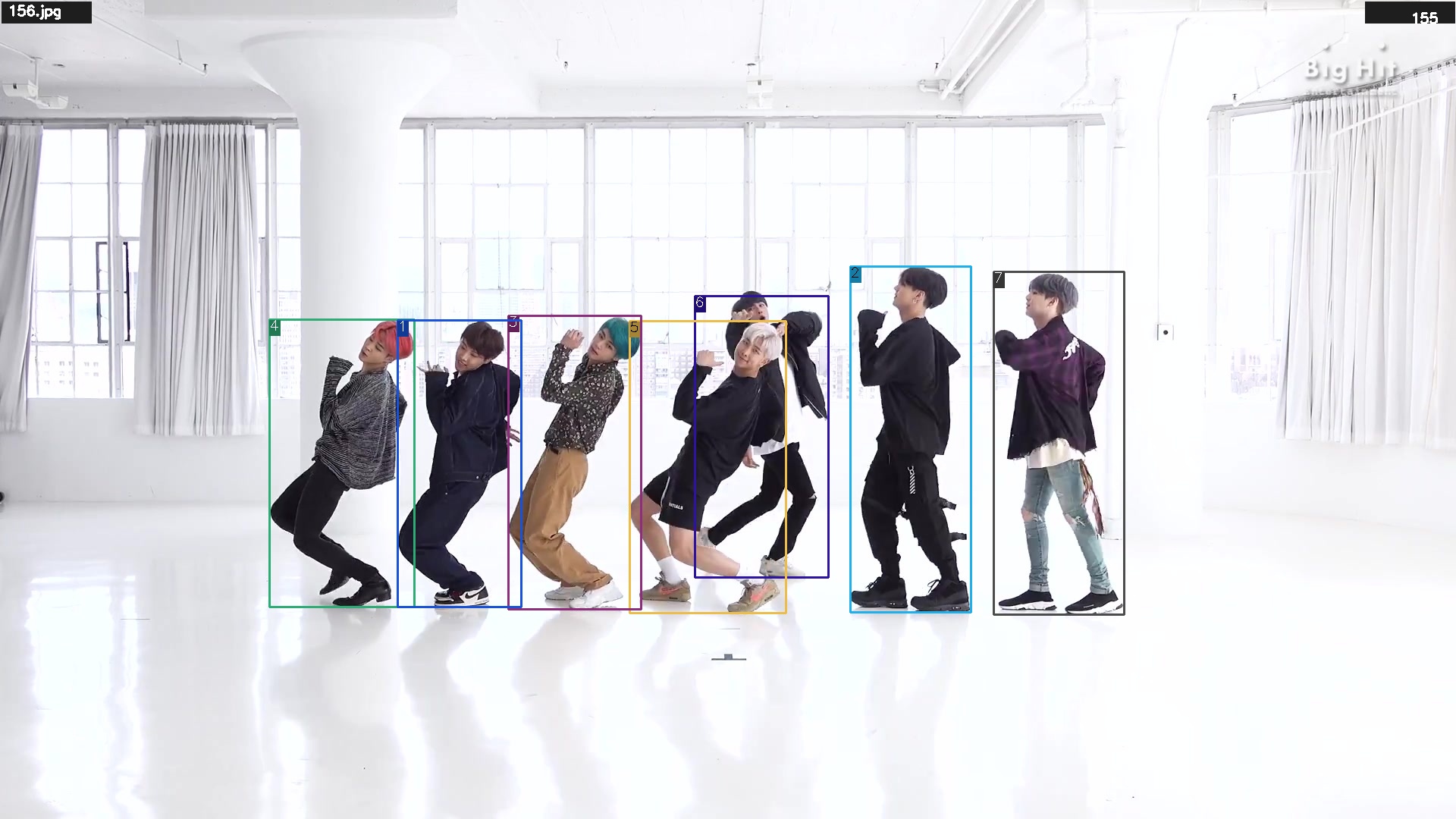} 
		\caption{The results through Hybrid-SORT.}
		\label{fig:vis2-a}
	\end{subfigure}
	\hfill
	\begin{subfigure}{0.48\linewidth}
		\includegraphics[width=\linewidth]{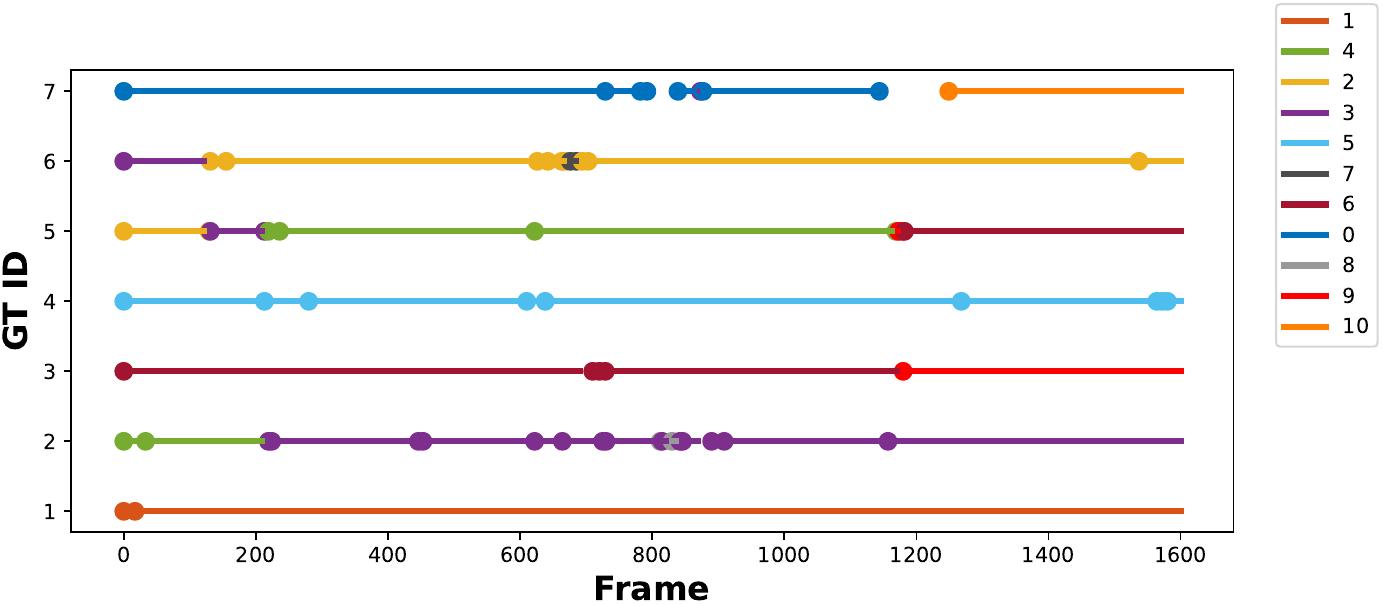}
		\includegraphics[width=\linewidth]{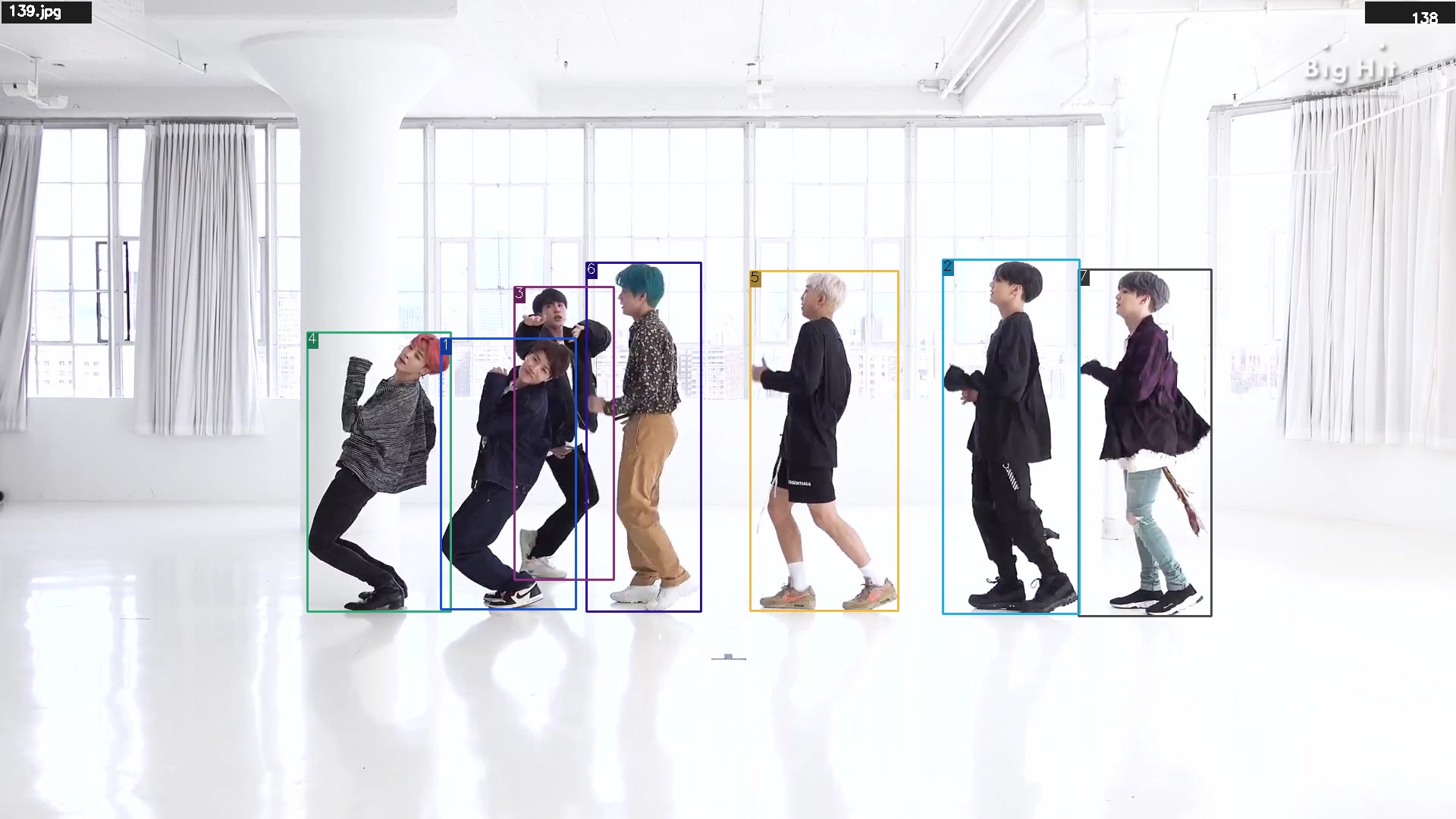} \\
		\includegraphics[width=\linewidth]{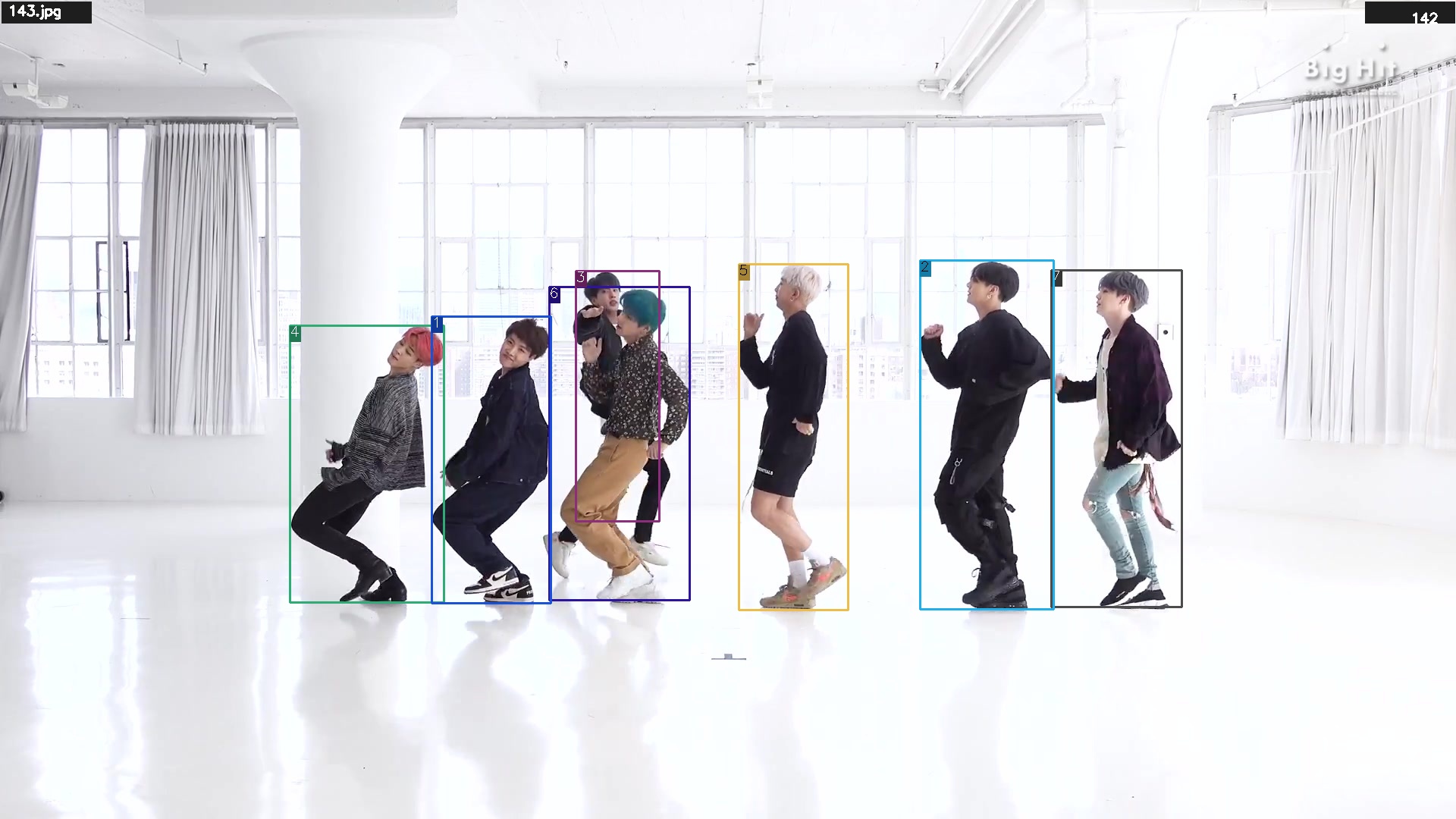} \\
		\includegraphics[width=\linewidth]{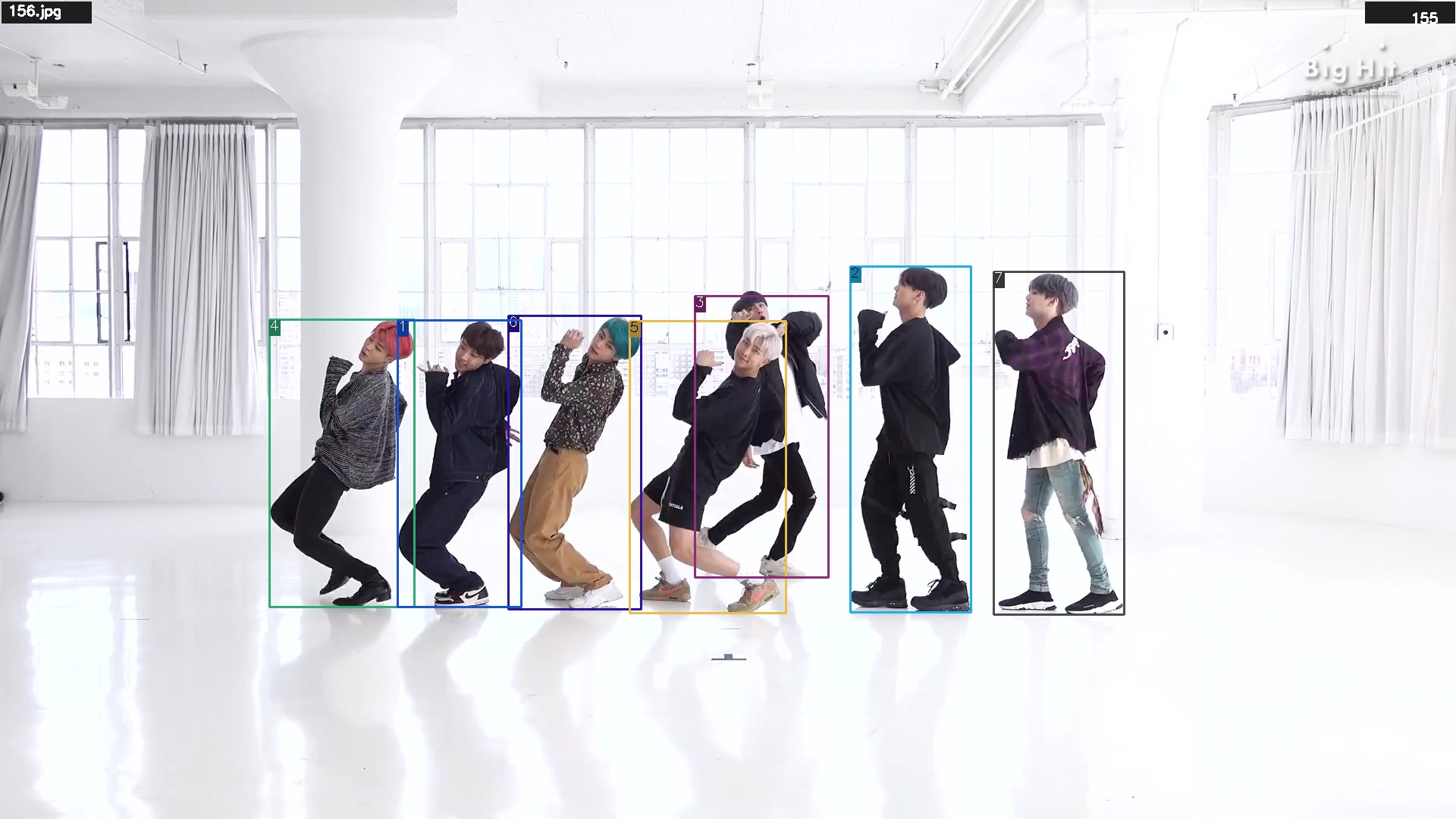}
		\caption{The results through OA-SORT.}
		\label{fig:vis2-b}
	\end{subfigure}
	\caption{The results on DanceTrack0058. The different color indicates different trajectory number and the dots on the line represent the starting point of trajectory interruption or identity change. GT ID represents the actual ID. The number in the legend represents the trajectory number rather than GT ID. The legend also reflects the total number of trajectories in the sequence.}
	\label{fig:vis2}
\end{figure*}

\begin{figure*}[t]
	\centering
	\begin{subfigure}{0.48\linewidth}
		\includegraphics[width=\linewidth]{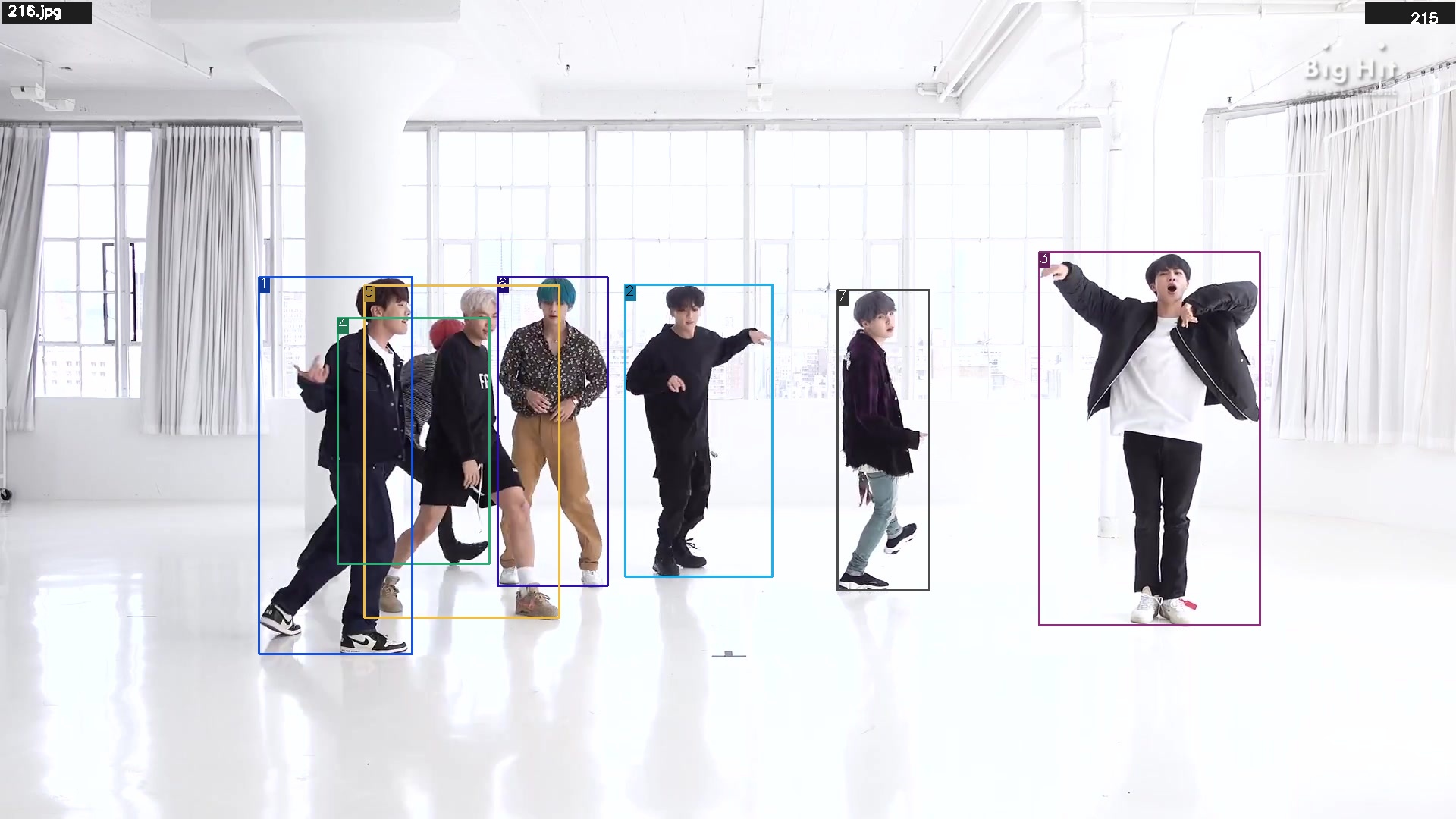} 
		\caption{\textbf{Frame} 215}
	\end{subfigure}
	\hfill
	\begin{subfigure}{0.48\linewidth}
		\includegraphics[width=\linewidth]{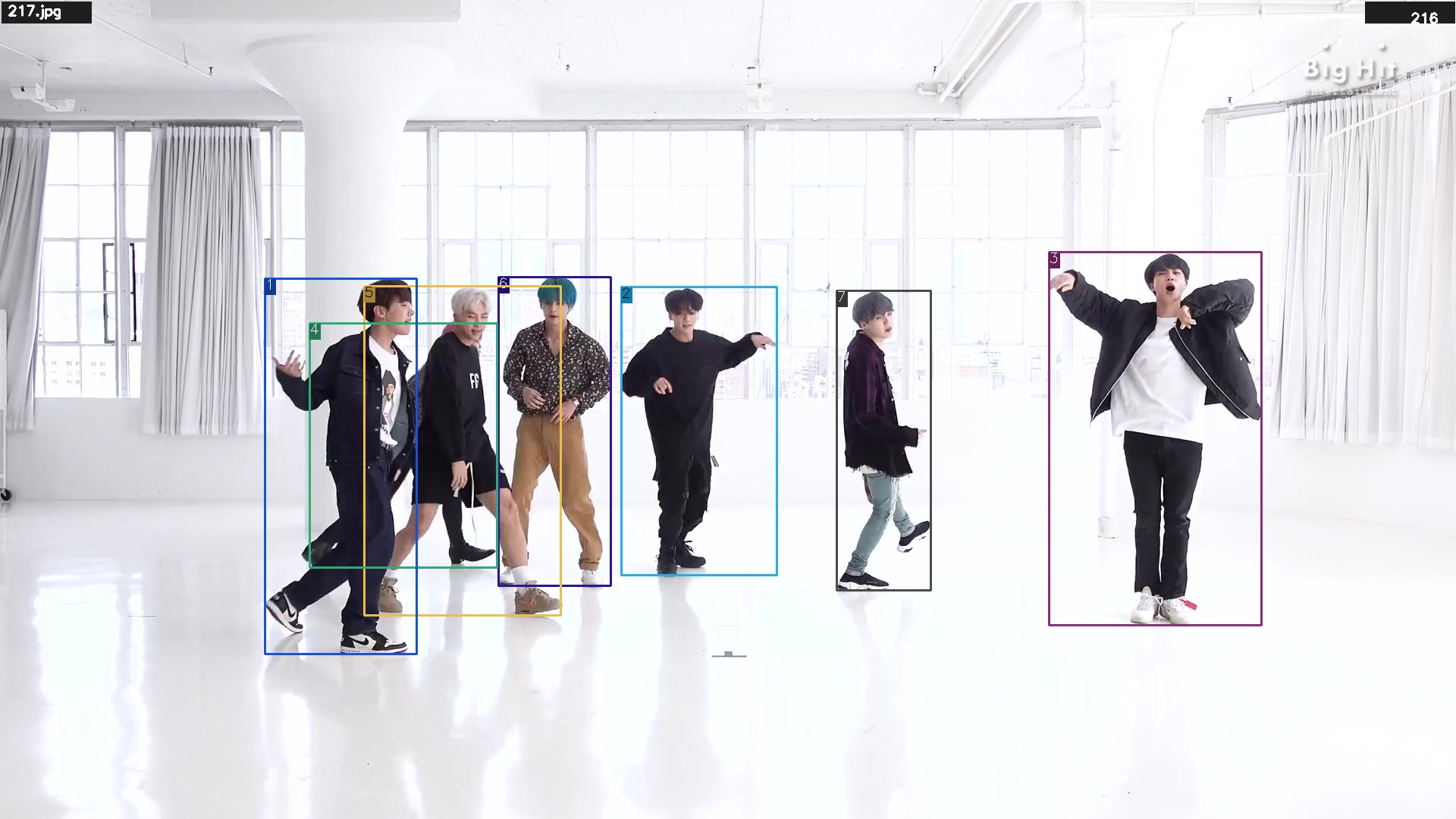}
		\caption{\textbf{Frame} 216}
	\end{subfigure}
	\\
	\begin{subfigure}{0.48\linewidth}
		\includegraphics[width=\linewidth]{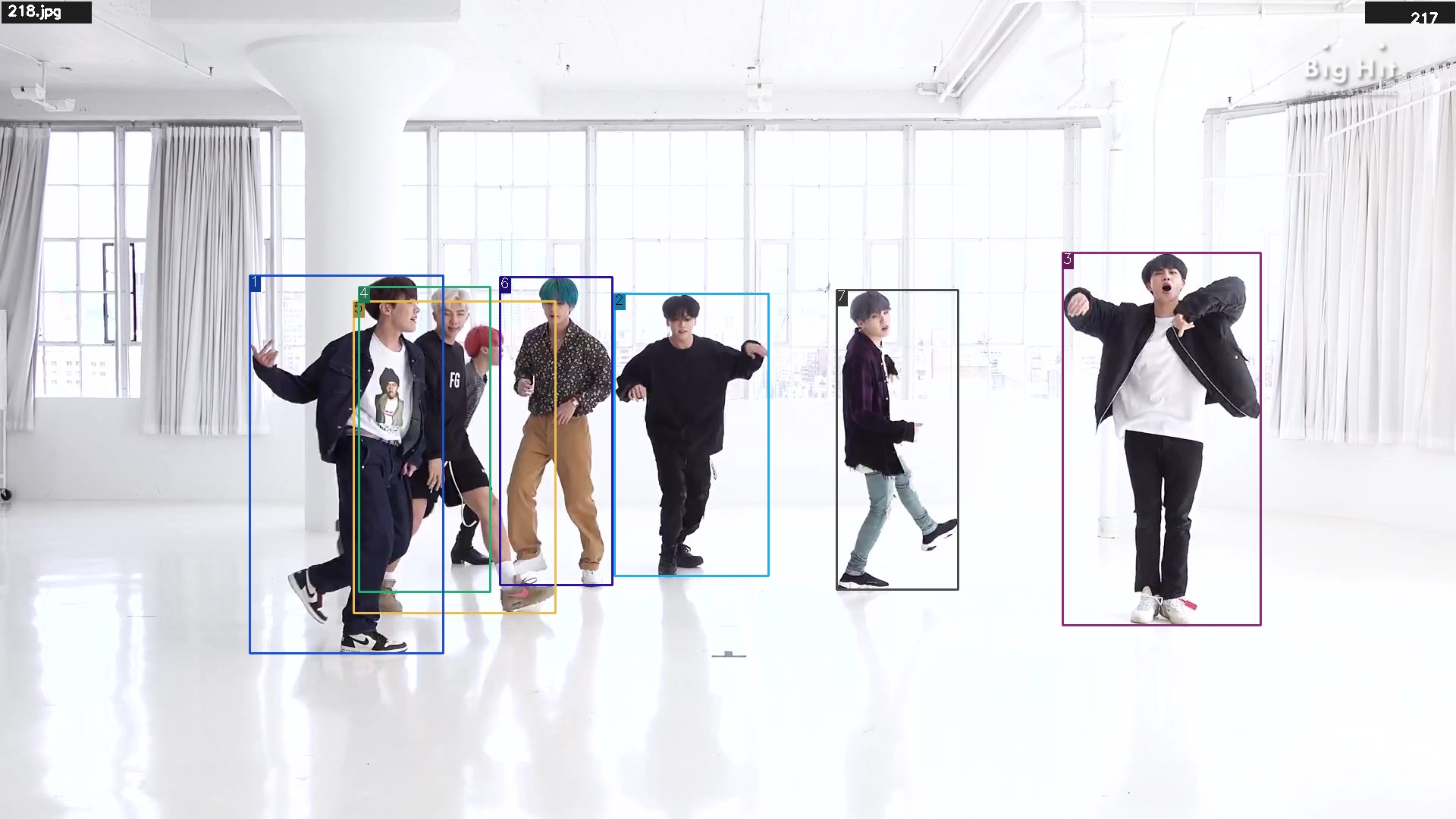}
		\caption{\textbf{Frame} 217}
	\end{subfigure}
	\hfill
	\begin{subfigure}{0.48\linewidth}
		\includegraphics[width=\linewidth]{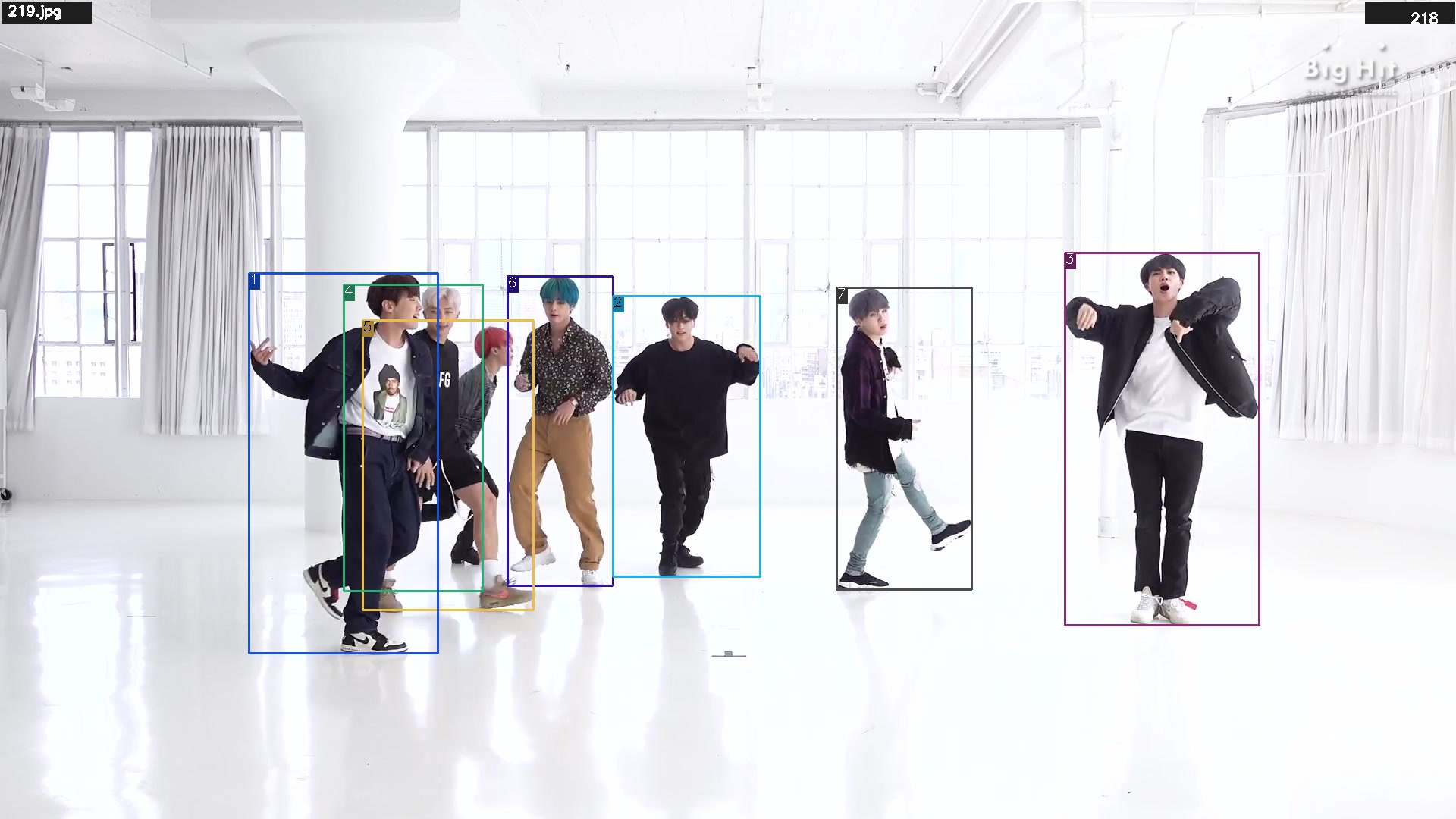}
		\caption{\textbf{Frame} 218}
	\end{subfigure}
	\caption{The failure case on DanceTrack0058. In Frame 217, due to the inaccurate detections of \#4 and \#5, the occlusion processing failed.}
	\label{fig: failure_case}
\end{figure*}

We analyze a video segment from the DanceTrack0005 under Hybrid-SORT and OA-SORT, as shown in \cref{fig:vis} and \cref{fig:vis2}. In frame 66 (in \cref{fig:vis}), Hybrid-SORT fails to maintain IDs of \#3 and \#2 due to position cost confusion brought by occlusion. In contrast, OA-SORT can stable their IDs utilizing OAO. However, occlusion-aware framework is difficult to handle severe inaccurate detection under strong competition, as shown for bounding box \#4 and \#5 in the \cref{fig: failure_case}. 

\end{document}